\pdfoutput=1
\documentclass[11pt]{article}

\usepackage[]{acl}

\usepackage{times}
\usepackage{latexsym}

\usepackage[T1]{fontenc}
\usepackage[utf8]{inputenc}

\usepackage{subcaption}
\usepackage{microtype}
\usepackage{url}
\usepackage{tikz,lipsum}
\usepackage[most]{tcolorbox}

\usepackage{amsfonts}
\usepackage{amsthm}
\newtheorem{theorem}{Theorem}

\usepackage{xcolor}
\usepackage[linesnumbered,ruled,vlined]{algorithm2e}
\usepackage[fencedCode,inlineFootnotes,citations,definitionLists,hashEnumerators,smartEllipses,hybrid]{markdown}

\SetCommentSty{mycommfont}

\SetKwInput{KwInput}{Input}                % Set the Input
\SetKwInput{KwOutput}{Output}              % set the Output

\usepackage{booktabs}
\usepackage{adjustbox}
\usepackage[english]{babel}
\usepackage{cleveref}
\usepackage{amsthm}
\usepackage{multirow}

\theoremstyle{definition}
\newtheorem{definition}{Definition}

\newcommand{\AdvNoise}{\texttt{FEDERATE}\xspace}
\newcommand{\AdvMultiple}{\texttt{Adversarial + Multiple}\xspace}
\newcommand{\Adv}{\texttt{Adversarial}\xspace}
\newcommand{\Uncon}{\texttt{Unconstrained}\xspace}

\usepackage{color}% Include colors for document elements
\usepackage[normalem]{ulem}
\title{Fair NLP Models with Differentially Private Text Encoders}

\author{Gaurav Maheshwari, Pascal Denis, Mikaela Keller,  Aur\'elien Bellet \\
Univ. Lille, Inria, CNRS, Centrale Lille, UMR 9189 - CRIStAL, F-59000 Lille, France\\
  \texttt{first\_name.last\_name@inria.fr} }

\begin{document}
\maketitle
\begin{abstract}
Encoded text representations often capture sensitive attributes about individuals (e.g., race or gender), which raise privacy concerns and can make downstream models unfair to certain groups. In this work, we propose \AdvNoise, an approach that combines ideas from differential privacy and adversarial training to learn private text representations which also induces fairer models. We empirically evaluate the trade-off between the privacy of the representations and the fairness and accuracy of the downstream model on four NLP datasets. Our results show that \AdvNoise consistently improves upon previous methods, and thus suggest that privacy and fairness can positively reinforce each other.
\end{abstract}

% Encoded text representations often capture sensitive attributes about individuals (e.g., gender, race, or age), which can raise privacy concerns and contribute to making downstream models unfair to certain groups. In this work, we propose FEDERATE, an approach that combines ideas from differential privacy and adversarial learning to learn private text representations which also induces fairer models. We empirically evaluate the trade-off between the privacy of the representations and the fairness and accuracy of the downstream model on two challenging NLP tasks. Our results show that FEDERATE consistently improves upon previous methods.

\section{Introduction}

% temp remove
% These instructions are for authors submitting papers to *ACL conferences using \LaTeX. They are not self-contained. All authors must follow the general instructions for *ACL proceedings,\footnote{\url{http://acl-org.github.io/ACLPUB/formatting.html}} and this document contains additional instructions for the \LaTeX{} style files.

% The rise of algorithmically driven decision-making systems trained on consumer data has raised several fairness and bias concerns. On the one hand, it is imperative to be non-discriminatory against a subgroup while making a decision. On the other hand, numerous laws have been proposed, such as GDPR~\cite{}, which restricts the use and revealing of sensitive information like gender or race of a person. These concerns have prompted various methods and tools in the domain of privacy~\cite{} and fairness~\cite{}, which are often oblivious to one another. 

% However, apart from societal concerns and laws forcing us to consider them together, several studies have shown that just enforcing one of these axes (privacy or fairness) can aggravate the problems on the other axis.  For instance,~\citet{} found that enforcing privacy can lead to accuracy differences in the subgroups, while~\citet{} showed that fair models could leak more training data of unprivileged groups. Thus, it is imperative to enforce fairness and privacy simultaneously.

Algorithmically-driven decision-making systems raise fairness concerns~\cite{DBLP:conf/fat/RaghavanBKL20, DBLP:conf/icis/BroekSH19}
% jagtiani2019roles
as they can be discriminatory against specific groups of people.
% In order to address these issues, various tools and mechanisms have been implemented based on fairness criteria such as demographic parity or equal opportunity.
% On the other hand,
% Apart from fairness issues,
%On the other hand, t
These systems have also been shown to leak sensitive information about the data of individuals used for training or inference, and thus pose privacy risks \cite{DBLP:conf/sp/ShokriSSS17}. % provide their data for training or inference.
% in the form of training data instances or sensitive attributes about these instances. This contradicts laws such as GDPR, which restricts the use and revealing of sensitive information like the gender or race of a person. These are issues tackled by the privacy-preserving machine learning community that develop solutions such as differential privacy.
Societal pressure as well as recent regulations % like GDPR
push for enforcing both privacy and fairness in real-world deployments, which is challenging as these notions are multi-faceted concepts that need to be tailored to the context. 
%Furthermore, privacy and fairness can be at odds with one another. For instance, r
Moreover, privacy and fairness can be at odds with one another: recent studies have shown that preventing a model from leaking information about its training data negatively impacts the fairness of the model and vice versa
\citep{DBLP:conf/nips/BagdasaryanPS19,DBLP:conf/fat/PujolMKHMM20,DBLP:conf/um/CummingsGKM19,DBLP:journals/corr/abs-2011-03731}.
% The issue of privacy and fairness does not exist in isolation. Apart from societal concerns and laws forcing us to consider them together, several studies have shown that just enforcing one of these axes (privacy or fairness) can aggravate the problems on the other axis.  For instance,~\citet{} found that enforcing privacy can lead to accuracy differences in the subgroups, while~\citet{} showed that fair models could leak more training data of unprivileged groups. Thus, it is imperative to enforce fairness and privacy simultaneously. 

% Like various definitions of fairness, privacy can be applied and studied in various contexts. Most of the above studies have focused on providing privacy over the training data, i.e., guarding against the membership inferencing attacks. However,
This paper studies fairness and privacy and their interplay in the NLP context, where these two notions have often been considered independently from one another. Modern NLP heavily relies on learning or fine-tuning encoded representations of text%, typically obtained as intermediate representations of a machine learning model
. Unfortunately, such representations often
% In NLP, the encoded representation of the underlying text by a model may
leak sensitive attributes (e.g., gender, race, or age) present explicitly or implicitly in the input text, even when such attributes are known to be irrelevant to the task~\cite{DBLP:conf/ccs/SongR20}. Moreover, the presence of such information in the representations may lead to unfair downstream models, as has been shown on various NLP tasks such as occupation prediction from text bios~\cite{DBLP:conf/fat/De-ArteagaRWCBC19}, coreference resolution~\cite{DBLP:conf/naacl/ZhaoWYOC18}, or sentiment analysis~\cite{kiritchenko-mohammad-2018-examining}. 

%. For instance, even after scrubbing explicit gender indicators from text such as names and pronouns,~\citet{DBLP:conf/fat/De-ArteagaRWCBC19} found that occupation prediction models still show a large correlation between accuracy and gender, indicating the use of implicit gender information.~\citet{DBLP:conf/naacl/ZhaoWYOC18} and~\citet{kiritchenko-mohammad-2018-examining} observed a similar phenomenon in coreference resolution and sentiment analysis, respectively.

Privatizing encoded representations is thus an important, yet challenging problem for which existing approaches based on subspace projection \citep{DBLP:conf/nips/BolukbasiCZSK16, DBLP:conf/acl/WangLRMOX20,karve-etal-2019-conceptor,DBLP:conf/acl/RavfogelEGTG20} or adversarial learning \citep{li-etal-2018-towards,DBLP:conf/emnlp/CoavouxNC18,DBLP:conf/eacl/HanBC21} do not provide a satisfactory solution. In particular, these methods lack any formal privacy guarantee, and it has been shown that an adversary can still recover sensitive attributes from the resulting representations with high accuracy~\cite{DBLP:conf/emnlp/ElazarG18,gonen-goldberg-2019-lipstick}.

% Recently, works such as~\citet{DBLP:conf/emnlp/LyuHL20,plant2021cape} have proposed using differential privacy (DP)
% , a mathematical definition of privacy which comes with rigorous guarantees \cite{DBLP:journals/fttcs/DworkR14}, 
% based text encoder in order to prevent sensitive attribute leakage.

Instead of relying on adversarial learning to prevent attribute leakage, \citet{DBLP:conf/emnlp/LyuHL20,plant2021cape} recently propose to add random noise to text representations so as to satisfy differential privacy (DP), a mathematical definition which comes with rigorous guarantees \citep{DBLP:conf/tcc/DworkMNS06}.
However, we uncover a critical error in their privacy analysis which drastically weakens their privacy claims. Moreover, their approach harms accuracy and fairness compared to adversarial learning.

In this work, we propose a novel approach (called \AdvNoise) to learn private text representations and fair models by combining ideas from DP with an adversarial training mechanism.
% \gmrm{In this work, we propose a novel approach (called \AdvNoise) to learn private text representations by combining ideas from differential privacy (DP), a mathematical definition of privacy which comes with rigorous guarantees \cite{DBLP:journals/fttcs/DworkR14}, with an adversarial training mechanism.}
More specifically, we propose a flexible end-to-end architecture in which (i) the output of an arbitrary text encoder is normalized and perturbed using random noise to make the resulting encoder differentially private, and (ii) on top of the encoder, we combine a classifier branch with an adversarial branch to actively induce fairness, improve accuracy and further hide specific sensitive attributes.
% We normalize and add calibrate noise to the encoded representation before applying adversarial training over the encoder.
% \mkrm{This architecture is trained end-to-end and can accommodate any type of text encoder while ensuring formal DP guarantees for the resulting text representations.}
% This is in contrast to recent attempts at using DP in NLP \citep{DBLP:conf/emnlp/LyuHL20,plant2021cape}, for which we uncover a critical error in the privacy analysis.
% We also propose novel stopping mechanisms, which leads to a more fair model without significant loss of accuracy.
% Moreover, we find an error in the sensitivity analysis in previously published works that employed DP in the context of NLP, leading to significantly less privacy guarantees than claimed. 

%%%%THIS IS AN ALTERNATIVE TO THE ABOVE PARAGRAPH%%%%%%%%%%%%%%%%
% To accomplish this, several methods have employed adversarial-based training mechanisms. However, adversarial training does not provide any privacy guarantees. Moreover, in practice, the adversarially robust model still leaks information. Instead, we propose to augment the adversarial method with a differential privacy-based randomization algorithm. We normalize and add calibrate noise to the encoded representation before applying an adversary over the encoder.  Our work provides theoretical privacy guarantees. We also propose novel stopping mechanisms, which leads to a more fair model without significant loss of accuracy. Moreover, we find an error related to sensitivity analysis in the previously published works which employed DP in the context of NLP. 

We empirically evaluate the privacy-fairness-accuracy trade-offs achieved by \AdvNoise over four datasets and find that it simultaneously leads to more private representations and fairer models than state-of-the-art methods while maintaining comparable accuracy.
% We also propose a new simple criterion for hyperparameter selection (applicable to all methods) which leads to significant improvements in fairness or privacy for a small cost in accuracy.
Beyond the superiority of our approach, our results bring valuable insights on the complementarity of DP and adversarial learning and the compatibility of privacy and fairness.
On the one hand, DP drastically reduces undesired leakage from adversarially trained representations, and has a stabilizing effect on the training dynamics of adversarial learning. % (which is notoriously unstable).
On the other hand, adversarial learning improves the accuracy and fairness of models trained over DP text representations.
Our main contributions are as follows:
\begin{itemize}
  \item We propose a new approach, \AdvNoise, which combines a DP encoder with adversarial learning to learn fair and accurate models from private representations.
  \item We identify and fix (with a formal proof) a critical mistake in the privacy analysis of previous work on learning DP text representations. 
% which nullify their privacy guarantees and render their experimental results unsubstantiated.  
%    \item Experiment evaluation over four varied dataset showcasing that \AdvNoise simultaneously leads to more private representations and fairer models than state-of-the-art methods while maintaining comparable accuracy.
  \item We empirically show that \AdvNoise leads to more private representations and fairer models than state-of-the-art methods while maintaining comparable accuracy. 
% \item \AdvNoise also provides better and smoother fairness-accuracy (resp. privacy-accuracy) trade-offs than purely adversarial (resp. purely noise-based) approaches on the large spectrum of possible trade-offs.
 \item Unlike previous studies, our empirical results suggest that privacy and fairness are compatible in our setting, and even mutually reinforce each other. 
\end{itemize}

The paper is organized as follows. Section~\ref{sec-prelims} provides background on differential privacy. Section~\ref{sec:approach} presents our approach. Section~\ref{sec:related-work} reviews related work. Experimental results and conclusions are given in Sections~\ref{sec:exp} and \ref{sec:conclu}.

%The paper is organized as follows. Section~\ref{sec-prelims} provides some useful background on differential privacy and Section~\ref{sec:approach} presents our approach. Section~\ref{sec:related-work} reviews some related work. We describe our experimental results in Section~\ref{sec:exp}, and conclude with final remarks in Section~\ref{sec:conclu}.

\section{Background: Differential Privacy}
\label{sec-prelims}

Differential Privacy (DP) \citep{DBLP:conf/tcc/DworkMNS06} provides a rigorous mathematical definition of the privacy leakage associated with an algorithm. It does not depend on assumptions about the attacker's capabilities and comes with a powerful algorithmic framework. For these reasons, it has become a de-facto standard in privacy currently used by the US Census Bureau~\cite{abowd2018us} and several big tech companies \citep{rappor:15,rappor2:16,telemetry:17}.
% computing keyboard statistics in iOS~\cite{2017LearningWP}.
This section gives a brief overview of DP, focusing on the aspects needed to understand our approach (see~\citet{DBLP:journals/fttcs/DworkR14} for an in-depth review of DP).

Over the last few years, two main models for DP have emerged:  
(i) Central DP (CDP)~\cite{DBLP:conf/tcc/DworkMNS06}, where raw user data is collected and processed by a trusted curator, which then releases the result of the computation to a third party or the public, and  
(ii) Local DP (LDP)~\cite{DBLP:journals/siamcomp/KasiviswanathanLNRS11} which removes the need for a trusted curator by having each user locally perturb their data before sharing it.
% where the data is perturbed locally by the user before releasing and thus needs no trust even at the curator level. 
% This work aims to create a training mechanism that provides privacy over the encoded user data. We envision a system where an encoding function and a mechanism to make these encoding private is provided to the end-user at the time of deployment. The user can thus "locally" encode and perturb the data before sending it to the curator for further analysis. 
Our work aims to create an encoder that leads to a private embedding of an input text, which can then be shared with an untrusted curator for learning or inference. We thus consider LDP, defined as follows.
% and give each user this encoding mechanism. 
\begin{definition}[Local Differential Privacy]
\label{def:dp}
A randomized algorithm $M:X\rightarrow O$ is  $\epsilon$-differentially private if for all pairs of inputs $x, x' \in X$ and all possible outputs $o \in O$:
\begin{equation}
    \Pr[M(x) = o] \leq e^{\epsilon}\Pr[ M(x') = o].
\end{equation}
\end{definition}

% Here $\epsilon$ controls the privacy with $\epsilon = 0$ implies complete privacy, whereas $\epsilon \rightarrow \infty$ gives no privacy guarantee.
\noindent LDP ensures that the probability of observing a particular output $o$ of $M$ should not depend too much on whether the input is $x$ or $x'$.
% changing the input of $M$ does not change too much the probability of observing any possible output.
The strength of privacy is controlled by $\epsilon$, which bounds the log-ratio of these probabilities for any $x,x'$. Setting $\epsilon = 0$ corresponds to perfect privacy, while $\epsilon \to \infty$ does not provide any privacy guarantees (as one may be able to uniquely associate an observed output to a particular input).
In our approach described in Section~\ref{sec:approach}, $x$ will be an input text and $M$ will be an encoding function which transforms $x$ into a private vector representation that can be safely shared with untrusted parties. 
% the ratio of probabilities is equal to $1$ implying complete privacy as $M$ is as likely to output $o$ for any pair of inputs. Whereas when $\epsilon \to \infty$, we have no privacy guarantees, as the probabilities diverge, meaning that it is more likely that a given input $\textbf{x}$ is associated to a given output $o$.
% Among other desirable properties, DP is robust to post-processing: any function $F$ applied over $M$ is still $\epsilon$-differential private. 

% A primary difference between CDP and LDP from a definition point of view is that input to $M$ in the case of CDP is a dataset while for LDP is a single data point.
\paragraph{Laplace  mechanism.} As clearly seen from Definition~\ref{def:dp}, an algorithm needs to be randomized to satisfy DP. % dp randomization algorithms ($M$) that provide DP perturb the input in some form. Several perturbation methods have been proposed in the literature, and the kind of input (model parameters, input dataset) they perturb gives privacy in different settings.
A classical approach to achieve $\epsilon$-DP for vector data is the Laplace mechanism~\cite{DBLP:conf/tcc/DworkMNS06}.
Given the desired privacy guarantee $\epsilon$ and an input vector $\textbf{x}\in\mathbb{R}^D$, this mechanism adds centered Laplace noise $\text{Lap}(\frac{\Delta}{\epsilon})$ independently to each dimension of $\textbf{x}$. The noise scale $\frac{\Delta}{\epsilon}$ is calibrated to $\epsilon$ and the \emph{L1-sensitivity} $\Delta$ of inputs:
% Given the desired privacy guarantee $\epsilon$ and an input vector $\textbf{x}\in\mathbb{R}^D$, this mechanism adds Laplace noise independently to each dimension in the input: 
%     \begin{equation}
%         \textbf{x}_{priv} \leftarrow \textbf{x} + \boldsymbol{\ell},
%         % \\ where l_i is \sim  \text{Lap}(\frac{\bigtriangleup f}{\epsilon})
%    \end{equation}
% where each entry of the vector $\boldsymbol{\ell}\in\mathbb{R}^D$ is sampled independently from a centered Laplace distribution with scale $\frac{\Delta}{\epsilon}$, denoted by $\text{Lap}(\frac{\Delta}{\epsilon})$. The noise scale is calibrated to $\epsilon$ and the L1-sensitivity $\Delta$ of the inputs defined as: 
\begin{equation}
\label{eq:sens}
    \Delta = \underset{\textbf{x}, \textbf{x}'\in X}{\max}~ % \sum_{i=1}^{D} |x_{i} - x'_{i}|.
    \|\textbf{x}- \textbf{x}'\|_1.
\end{equation}
% Here, $\textbf{x}$ and $\textbf{x'}$ are any two arbitrary encoded user input of $D$ dimension.
% Since the encoded inputs are normalized before before adding noise, the sensitivity in our case is 2. 
In our work, we propose an architecture in which the Laplace mechanism is applied on top of a trainable encoder to get private representations of input texts, and is further combined with adversarial training to learn fair models.

% The randomization algorithm includes methods like Laplace mechanism~\cite{}, and Gaussian mechanism\cite{}, which end users can employ to ensure privacy.  
% \paragraph{Post-Processing Property:} It is an important and widely used property of DP which states that any deterministic or random function $F$ applied over $M$ is still differential private. 
% It allows adversary to apply any sophisticated method over privatized encoding and still remain deferentially private.  
% Thus, this property enables us to apply adversarial mechanisms over $M$ and still enjoy all the implications of differential privacy.

\section{Approach}
\label{sec:approach}

% \begin{tcolorbox}[colback=red!5!white,colframe=red!75!black]
% \begin{itemize}
%     \item Problem setting
%     \item Approach 
%     \item After Adv step it is still private due to post-processing property of DP
% \end{itemize}
% \end{tcolorbox}

% \subsection{Problem Setting}

% his work aims to create a training mechanism that provides privacy over the encoded user data. We envision a system where an encoding function and a mechanism to make these encoding private is provided to the end-user at the time of deployment. The user can thus "locally" encode and perturb the data before sending it to the curator for further analysis. 

%This work aims to create a training mechanism that provides privacy over the encoded user data and ensures subsequent model fairness.
% In this work we aim to create a training mechanism which leads to privacy over model's representation and ensure subsequent model fairness. Specifically, our attack scenario is similar to~\citet{DBLP:conf/emnlp/CoavouxNC18}, where an eavesdropper or an untrusted server aims to infer some sensitive information $z$ from the latent representation of the user input.

We consider a scenario similar to \citet{DBLP:conf/emnlp/CoavouxNC18}, where a user locally encodes its input data (text) $x$ into an intermediate representation $E_{priv}(x)$ which is then shared with an untrusted curator to predict the label $y$ associated with $x$ using a classifier $C$. Additionally, an attacker (which may be the untrusted curator or an eavesdropper) may observe the intermediate representation $E_{priv}(x)$ and try to infer some sensitive (discrete) attribute $z$ about $x$ (e.g., gender, race etc.). Our goal is to learn an encoder $E_{priv}$ and classifier $C$ such that (i) the attacker performs poorly at inferring $z$ from $E_{priv}(x)$, (ii) the classifier $C(E_{priv}(x))$ is fair with respect to $z$ according to some fairness metric, and (iii) $C$ accurately predicts the label $y$.

\begin{figure}
    \centering
    \begin{minipage}{0.50\textwidth}
        \centering
        \includegraphics[width=0.9\textwidth]{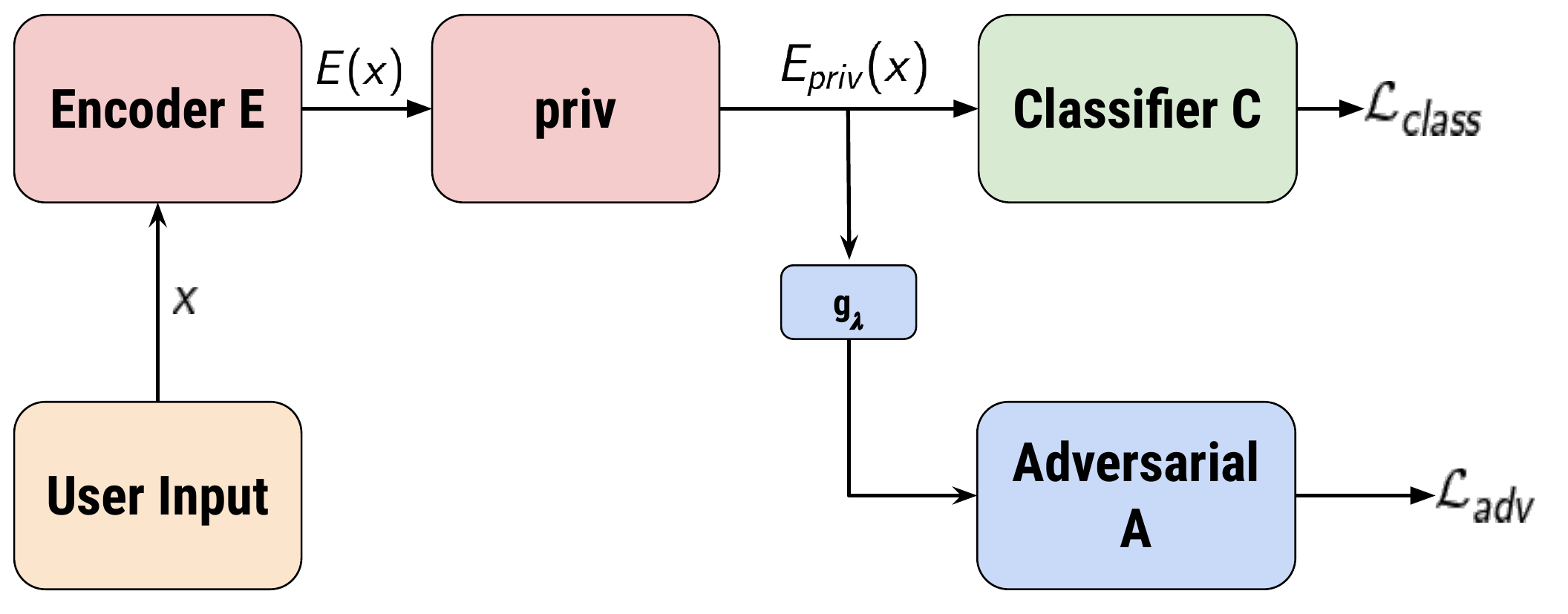} % first figure itself
        \caption{
Overview of our \AdvNoise approach. The text input $x$ is transformed to $E(x)\in\mathbb{R}^D$ by the text encoder $E$. The encoded input is then made private by the privacy layer $priv$, which involves normalization and addition of Laplace noise. The resulting private representation $E_{priv}(x)\in\mathbb{R}^D$ is then used by the main task classifier $C$. It also serves as input to the adversarial layer $A$ which is connected to the main branch via a radient reversal layer $g_{\lambda}$. The light red boxes represent the Differentially Private Encoder (Sec.~\ref{sec:dp_encoder}), and the light blue boxes represent the Adversarial component (Sec.~\ref{sec:adv_component}). }
\label{arch-overview}
    \end{minipage}\hfill 
        
\end{figure}

% \gmrp{
To achieve the above goals we introduce \AdvNoise (for Fair modEls with DiffERentiAlly private Text Encoders), which combines
% \abrm{ideas from DP and adversarial learning by integrating a randomized mapping into the encoder and modeling the adversary in the training phase to improve the fairness of the classifier.}
two components: a differentially private encoder and an adversarial branch. Figure~\ref{arch-overview} shows an overview of our proposed architecture. 
\subsection{Differentially Private Encoder}
\label{sec:dp_encoder}

% \paragraph{Proposed architecture.}
We propose a generic private encoder construction $E_{priv}=priv\circ E$ composed of two main components. The first component $E$ can be any encoder which maps the text input to some vector space of dimension $D$. It can be a pre-trained language model along with a few trainable layers, or it can be trained from scratch. The second component $priv$ is a randomized mapping which transforms the encoded input to a differentially private representation. Given the desired privacy guarantee $\epsilon>0$, this mapping is obtained by applying 
the Laplace mechanism (see Section~\ref{sec-prelims}) to a normalized version of the encoded representation $E(x)$:
% as function $priv$. This involves normalizing the representation and then adding coordinate-wise laplacian noise. The transformation can be sumarized as: 
%     \begin{align}
% \textbf{x} &\leftarrow \textbf{x} / \|\textbf{x}\|_1 \\
%         \textbf{x}_{priv} &\leftarrow \textbf{x} +  (l_1, l_2, \ldots, l_D) , l_i \sim  \text{Lap}(\frac{\bigtriangleup f}{\epsilon})
%     \end{align}
    \begin{equation}
    \label{eq:our_lpriv}
        priv(E(x)) = E(x)/\|E(x)\|_1 + \boldsymbol{\ell},
    \end{equation}
where each entry of $\boldsymbol{\ell}\in\mathbb{R}^D$ is sampled independently from $\text{Lap}(\frac{2}{\epsilon})$. We will prove that  $E_{priv}=priv\circ E$ satisfies $\epsilon$-DP in Section~\ref{sec:privacy_analysis}.

\subsection{Adversarial Component}
\label{sec:adv_component}

To improve the fairness of the downstream classifier $C$, we model the adversary by another classifier $A$ which aims to predict $z$ from the privately encoded input $E_{priv}(x)$. The encoder $E_{priv}$ is optimized to fool $A$ while maximizing the accuracy of the downstream classifier $C$. Specifically, given $\lambda> 0$, we train $E_{priv}$, $C$ and $A$ (parameterized by $\theta_E$, $\theta_C$, and $\theta_A$ respectively) to optimize the following objective:
% \begin{equation}
%     \underset{enc, c, adv}{\mathrm{argmin}}\ L(c(M(enc(x))),y) +  L(adv(g_{\lambda}(M(enc(x)))),z)
% \end{equation}
%\begin{align*}
%    \hat{\theta}_{E_{priv}}, \hat{\theta}_{C} = & \underset{\theta_{E_{priv}}, \theta_C}{\mathrm{argmin}} E(\theta_{E_{priv}}, \theta_C, \hat{\theta}_A) \\
%    \hat{\theta}_{A} =  & \underset{\theta_A}{\mathrm{argmax}} E(\hat{\theta}_{E_{priv}}, \hat{\theta_C}, \theta_A)
%\end{align*}
\begin{equation}
\label{eq:obj}
      \underset{\theta_E, \theta_C}{\mathrm{min}} \underset{\theta_A}{\mathrm{
      max}}\> {\cal L}_{class}(\theta_E, \theta_C) - \lambda 
      {\cal L}_{adv}(\theta_E, \theta_A),
\end{equation}
where ${\cal L}_{class}(\theta_E, \theta_C)$ is the cross-entropy loss for the $C\circ E_{priv}$ branch and ${\cal L}_{adv}(\theta_E, \theta_A)$ is the cross-entropy loss for the $A\circ E_{priv}$ branch.

\subsection{Training}
\label{sec:train}

% \paragraph{Training phase.}
We train the private encoder $E_{priv}$ and the classifier $C$ from a set of public tuples $(x,y,z)$
% so as to guard against the adversary with access to $\textbf{x}_{priv}$ and have the predictions be fair on some fairness metric. 
% (i) An encoder or a feature extractor, enc, which encodes user input ,
% (ii) a DP transformation function M which transforms encoded text to private representation, and
% (iii) a classifier c which takes this transformed representation and predicts the label y.
% Here the adversary has access to private representation through which it tries to infer sensitive attributes. 
% \subsection{Methodology}
% In order to achieve the above goals,
by optimizing \eqref{eq:obj} with backpropagation using a gradient reversal layer $g_\lambda$~\cite{pmlr-v37-ganin15}. The latter acts like an identity function in the forward pass but scales the gradients passed through it by $-\lambda$ in the backward pass. This results in $E_{priv}$ receiving opposite gradients to $A$. We give pseudo-code in Appendix~\ref{sec:algorithm}.
%${\cal L}$ represents the loss function, which in our case is cross-entropy and $y$ is the actual label of the task.
% Here, $enc_{priv}$ will be $priv(enc(\cdot))$.

% define function M 
% give the algorithm 

% \paragraph{Inference phase.} Once trained, $E_{priv}$ can be used to privately encode new data points which can then be fed into the classifier $C$ for inference. Note that by the post-processing property of DP, applying $C$ or any other function on top to $E_{priv}$ preserves the $\epsilon$-DP guarantee of $E_{priv}$.
% In our experiments, we will empirically evaluate the privacy of $E_{priv}(\cdot)$ and the fairness of $C(E_{priv}(\cdot))$ and show that our approach consistently provides better privacy-fairness-accuracy trade-offs than previous methods.

\subsection{Privacy Analysis}
\label{sec:privacy_analysis}

We show the following privacy guarantee.
% \abrm{As
% the L1 normalization in \eqref{eq:our_lpriv} ensures that
% the L1 sensitivity of the \emph{normalized} representation is bounded by $2$ for any $E$, $E_{priv}=priv\circ E$ is $\epsilon$-DP.}
% Here $\bigtriangleup f$ is the L1-sensitivity of the extracted representation, and $\epsilon$ is the privacy budget. Since the encoded inputs are normalized before before adding noise, the sensitivity in our case is 2.

\begin{theorem}
\label{thm:priv}
Our encoder $E_{priv}$ and the downstream predictions $C\circ E_{priv}$ satisfy $\epsilon$-DP.
\end{theorem}

The proof is given in Appendix~\ref{app:privacy_proof}. Theorem~\ref{thm:priv} shows that the encoded representations produced by $E_{priv}$ have provable privacy guarantees: in particular, it bounds the risk that the sensitive attribute $z$ of a text $x$ is leaked by $E_{priv}(x)$.\footnote{More generally, the DP guarantee bounds the risk that \emph{any} attribute of $x$ is leaked through $E_{priv}(x)$.} These privacy guarantees naturally extend to the downstream prediction $C(E_{priv}(x))$ due to the post-processing properties of DP (see Appendix~\ref{app:privacy_proof} for details).
%In our experiments, we will complement these analytical privacy guarantees through empirical metrics, and show that our approach consistently provides better privacy-fairness-accuracy trade-offs than previous methods.

\paragraph{Error in previous work.} We found a critical error in the privacy analysis of previous work on differential private text encoders \citep{DBLP:conf/emnlp/LyuHL20,plant2021cape}. In a nutshell, they incorrectly state that normalizing each entry of the encoded representation in $
[0,1]$ allows to bound the sensitivity of their representation by 1, while it can in fact be as large as $D$ (the dimension of the representation). As a result, the privacy guarantees are dramatically weaker than what the authors claim: the $\epsilon$ values they report should be multiplied by $D$. In contrast, the L1 normalization we use in \eqref{eq:our_lpriv} ensures that the sensitivity of $E$ is bounded by $2$. We provide more details in Appendix~\ref{sec:correction}. 

Interestingly,~\citet{DBLP:conf/emnlp/Habernal21} recently identified an error in ADePT~\cite{krishna-etal-2021-adept}, a differentially private auto-encoder for text rewriting. However, the error in ADePT is different from the one in \citet{DBLP:conf/emnlp/LyuHL20,plant2021cape}: the problem with ADePT is that it calibrates the noise to L2 sensitivity, while the Laplace mechanism requires L1 sensitivity. %Whereas in \citet{DBLP:conf/emnlp/LyuHL20,plant2021cape}, it is the incorrect bound on L1 sensitivity that causes the privacy error. 
These errors call for greater scrutiny of differential privacy-based approaches in NLP---our work contributes to this goal.

\section{Related Work}
\label{sec:related-work}

% In this section, we begin by giving a brief overview of the fairness and privacy approaches commonly employed in NLP literature. We then describe studies that tackle these issues together and finally an approach which is closest to our setting.

% In this section, we begin by giving a brief overview of the fairness and privacy approaches commonly employed in NLP literature. We then describe studies that tackle these issues together and, finally, an approach closest to our setting.
% This section reviews related lines of work, highlighting the main differences with our approach.

% One of the most common approaches to induce a fair model or private representations, i.e., representations that do not leak sensitive information in NLP, is to employ an adversarial-based training mechanism.
\paragraph{Adversarial learning.} In order to improve model fairness or to prevent leaking sensitive attributes, several approaches employ adversarial-based training.
% The general idea \citep{adv_learning} is to train an intermediate representation so as to maximize the prediction accuracy for the downstream task of interest while minimizing the performance of an adversary seeking to recover sensitive attributes from the representation (as modeled by an adversarial branch).
For instance,~\citet{li-etal-2018-towards} propose to use a different adversary for each protected attribute, while~\citet{DBLP:conf/emnlp/CoavouxNC18} consider additional loss components to improve the privacy-accuracy trade-off of the learned representation. \citet{DBLP:conf/eacl/HanBC21} introduce multiple adversaries focusing on different aspects of the representation by encouraging orthogonality between pairs of adversaries.
% Their loss function encourages orthogonality between pairs of adversaries and leads to some improvements in the fairness of downstream models at the cost of higher training complexity.
% On similar lines,~\cite{} proposes an adversarial scrubbing mechanism to remove sensitive information from the encoder. However, they purely focus on information leakage, and not on  fairness. Moreover, 
Recently,~\citet{DBLP:conf/emnlp/ChowdhuryGLOSC21} propose an adversarial scrubbing mechanism. However, they purely focus on information leakage, and not on fairness. Moreover, unlike our approach, these methods do not offer formal privacy guarantees.
% Unlike our approach, these methods do not offer formal privacy guarantees.
In fact, it has been observed that one can recover the sensitive attributes from % do not properly remove the sensitive information from the intermediate representation. For instance,
% though with several methods relying on adversarial training, their success is  limited as they do not provide any guarantees related to privacy. Moreover, ~
the representations by training a post-hoc non linear classifier \citep{DBLP:conf/emnlp/ElazarG18}. This is confirmed by our empirical results in Section~\ref{sec:exp}.
% Further, these methods do not provide any guarantees related to privacy.

\paragraph{Sub-space projection.} %Parallel to these approaches,
A related line of work focuses on debiasing text representations using projection methods \cite{DBLP:conf/nips/BolukbasiCZSK16, DBLP:conf/acl/WangLRMOX20,karve-etal-2019-conceptor}. The general approach involves identifying and removing a sub-space associated with sensitive attributes.
% and removing this subspace from the neutral words.
% \mkrm{A key advantage over adversarial learning is that these methods do not use a task-specific loss. They instead rely on a manual selection of words in the vocabulary to estimate the sensitive sub-space, making them difficult to generalize to new attributes.}
However, they rely on a manual selection of words in the vocabulary
% to estimate the sensitive sub-space,
which is difficult to generalize to new attributes.
% them difficult to generalize.
Furthermore, \citet{gonen-goldberg-2019-lipstick}
% found bias to be deeply ingrained in these representations, and 
showed that sensitive attributes still remain present even after applying these approaches.
% Another disadvantage of these approaches is their reliance on hand-selecting words in the vocabulary to represent sensitive attributes making them difficult to generalize.

Recently,~\citet{DBLP:conf/acl/RavfogelEGTG20} propose Iterative Null space Projection (INLP). It involves iteratively training a linear classifier to predict sensitive attributes followed by projecting the representation on the classifier's null space. However, the method can only remove linear information from the representation. %As a result, a nonlinear adversary can still retrieve sensitive information. 
By leveraging DP, our approach provides robust guarantees that do not depend on the expressiveness of the adversary, thereby providing protection against a wider range of attacks.

\paragraph{DP and fairness.} Recent work has studied the interplay between DP and (group) fairness in the setting where one seeks to prevent a model from leaking information about individual training points. Empirically, this is evaluated through membership inference attacks, where an attacker uses the model to determine whether a given data point was in the training set \citep{DBLP:conf/sp/ShokriSSS17}. While \citet{DBLP:journals/corr/abs-1906-00389} observed that DP reduces disparate vulnerability to such attacks, it has also been shown that DP can exacerbate unfairness \cite{DBLP:conf/nips/BagdasaryanPS19,DBLP:conf/fat/PujolMKHMM20}. Conversely, \citet{DBLP:journals/corr/abs-2011-03731} showed that enforcing a fair model leads to more privacy leakage for the unprivileged group. This tension between DP and fairness is further confirmed by a formal incompatibility result between $\epsilon$-DP and fairness proved by \citet{DBLP:conf/um/CummingsGKM19}, albeit in a restrictive setting.
Some recent work attempts to train models under both DP and fairness constraints \citep{DBLP:conf/um/CummingsGKM19,DBLP:journals/corr/abs-2003-03699, liu2020fair}, but this typically comes at the cost of enforcing weaker privacy guarantees for some groups. Finally, \citet{DBLP:conf/icml/JagielskiKMORSU19} train a fair model under DP constraints only for the sensitive attribute.

A fundamental difference between this line of work and our approach lies in the kind of privacy we provide.
While the above approaches study (central) DP as a way to design algorithms which protect training points from membership inference attacks on the model, we construct a private encoder such that the encoded representation does not leak sensitive attributes of the input. % do not differ too much. 
% Instead of focusing on membership inference attacks, we focus on creating private encoded representation, i.e., once passed through the DP function,  any two representations would not differ significantly.
Thus, unlike previous work, we provide privacy guarantees with respect to the model's intermediate representation for data unseen at training time, and empirically observe that in this case
% privacy improves the fairness of models learned from such representations.
privacy and fairness are compatible and even mutually reinforce each other.
% However, we do not guarantee any privacy of the training set.
% Our setting is akin to the adversarial-based approaches described at the start of the section.
% We give more details on our setup and attack scenario in the approach section.

\paragraph{DP representations for NLP.} In a setting similar to ours, \citet{DBLP:conf/emnlp/LyuHL20} propose to use DP to privatize model's intermediate representation. Unlike their method, we actively promote fairness by using an adversarial training mechanism, which leads to more private representations and fairer models in practice. Importantly, we also uncover a critical error in their privacy analysis (see Sec.~\ref{sec:dp_encoder}).
% We also found a critical error in their privacy analysis, where they incorrectly bound the sensitivity of their representation by 1 while it can in fact be as large as $D$ (the dimension of the representation, which is typically in the hundreds). As a result, the privacy guarantees are significantly weaker than the authors claim: the $\epsilon$ values they report should be multiplied by $D$. We provide more details in Appendix~\ref{sec:correction}.

Concurrent to and independently from our work, \citet{plant2021cape} propose an adversarial-driven DP training mechanism. However, they do not consider fairness, whereas we focus on enforcing both fairness and privacy. Moreover, their method has the same incorrect analysis as \citet{DBLP:conf/emnlp/LyuHL20}.

\section{Experiments}
\label{sec:exp}
Recall that we are interested in approaches that are not only accurate but also fair and private at the same time. However, these three dimensions are not independent and are not straightforwardly amenable to a single evaluation metric. Thus, we present experiments aiming at (i) showcasing the privacy-fairness-accuracy tradeoffs of different approaches and then (ii) analyzing privacy-accuracy and fairness-accuracy tradeoffs separately. We begin by describing the datasets and the metrics.

% In this section, we present experiments aiming to (i) assess the privacy-fairness-accuracy trade-offs of different approaches and (ii) analyze privacy-accuracy and fairness-accuracy tradeoffs separately. 

% We begin by describing the setup, including datasets, baselines, and the fairness and privacy metrics we use. 
% Following the footsteps of~\citet{DBLP:conf/acl/RavfogelEGTG20}, we begin our experiments with a controlled setup and then evaluate the over dataset reflecting more real world bias.
% in a less artificial settings. 
% The objective of this experiment is to evaluate the privacy of $E_{priv}(\cdot)$ and the fairness of $C(E_{priv}(\cdot))$ at the same time and compare it with existing approaches. 

\begin{figure*}[t]
  \centering
  \includegraphics[width=\textwidth]{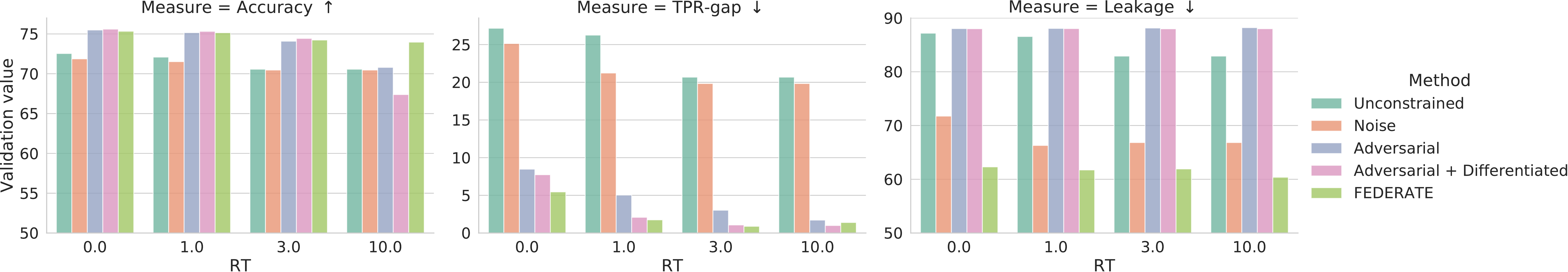}
  \caption{Validation accuracy, fairness and privacy of various approaches for different relaxation threshold (RT) (see Section~\ref{sec:relaxation}) on Twitter Sentiment. When RT is increased, we select models with potentially lower accuracy on the validation set but are more fair (lower TPR-gap).
  % We observe that fairness improves (TPR-gap decreases) with increasing RT for all approaches with little change in validation accuracy or privacy.
  Our approach \AdvNoise consistently achieves better accuracy-fairness-privacy trade-offs than its competitors across all RTs.}
  \label{fig:rt-tradeoff}
\end{figure*}

\paragraph{Datasets.}  % We employ Twitter Sentiment analysis and Bias in Bios~\cite{DBLP:conf/fat/De-ArteagaRWCBC19} dataset to illustrate various tradeoffs and performances of different approaches.

We consider 4 different datasets: (i) \textit{Twitter Sentiment} \citep{blodgett-etal-2016-demographic} consists of 200k tweets annotated with a binary sentiment label and a binary ``race'' attribute corresponding to African American English (AAE) vs. Standard American English (SAE) speakers; (ii) \textit{Bias in Bios} \cite{DBLP:conf/fat/De-ArteagaRWCBC19} consists of 393,423 textual biographies annotated with an occupation label (28 classes) and a binary gender attribute; (iii) \textit{CelebA}~\cite{DBLP:conf/iccv/LiuLWT15} is a binary classification dataset with a binary sensitive attribute (gender); (iv) \textit{Adult Income}~\cite{DBLP:conf/kdd/Kohavi96} consists of 48,842 instances with binary sensitive attribute (gender). Our setup for the first two dataset is similar to~\citet{DBLP:conf/acl/RavfogelEGTG20} and~\citet{DBLP:conf/eacl/HanBC21}. Appendix~\ref{appendix-dataset} provides detailed description of these datasets, including sizes, pre-processing, and the challenges they pose to privacy and fairness tasks. Due to lack of space, results and analyses for \textit{Adult Income} and \textit{CelebA dataset} are given in Appendix~\ref{sec-appendix-extended-evaluation}, but note that they exhibit similar trends. The preprocessed versions of the datasets can be downloaded from this anonymized URL.\footnote{\url{https://drive.google.com/uc?id=1ZmUE-g6FmzPPbZyw3EOki7z4bpzbKGWk}}

\paragraph{Fairness metrics.} For Twitter Sentiment we report the True Positive Rate Gap (TPR-gap), which measures the true positive rate difference between the two sensitive groups (gender/race) and is closely related to the notion of equal opportunity.
% We also report False Positive Rate Gap (FPR-gap), which, coupled with TPR-gap, corresponds to equalized odds~\cite{DBLP:conf/nips/HardtPNS16}.
Formally, denoting by $y\in\{0,1\}$ the ground truth binary label, $\hat{y}$ the predicted label and $z\in\{g,\neg g\}$ the sensitive attribute, TPR-gap is defined as:
\begin{equation*}
        \textrm{TPR-gap} = P_g(\hat{y}=1|y=1) - P_{\neg g}(\hat{y}=1|y=1).
\end{equation*}
% FPR-gap is defined similarly (see Appendix~\ref{sec:fpr}).

For Bias in Bios, which has 28 classes,
% makes it difficult to report  in a table. We thus
we follow~\citet{romanov-etal-2019-whats} and report the root mean square of TPR-gaps (GRMS) over all occupations $y \in O$ to obtain a single number:
\begin{equation}
    \text{GRMS} = \sqrt{(1/|O|)\textstyle\sum_{y \in O} (\textrm{TPR-gap}_y)^2 }.
\end{equation}

\paragraph{Privacy metrics.} 
% To measure the privacy of a text encoder, we use the accuracy of a two-layer adversarial network which predicts the sensitive attribute from the representation (Leakage). This classifier is trained on the validation set and evaluated on the test set. 

We report two metrics for privacy: (i) \textbf{Leakage}: the accuracy of a two-layer classifier which predicts the sensitive attribute from the encoded representation, and (ii) \textbf{Minimum Description Length (MDL)}~\cite{DBLP:conf/emnlp/VoitaT20}, which quantifies the amount of ``effort'' required by such a classifier to achieve a certain accuracy. A higher MDL means that it is more difficult to retrieve the sensitive attribute from the representation. The metric depends on the dataset and the representation dimension, and thus cannot be compared across different datasets.
% \begin{itemize}
%     \item \textbf{Leakage}: Accuracy of a two layer classifier which predicts the sensitive attribute from the encoded representation. 
%     % It is trained on the validation set and evaluated on the test set. 
%     \item \textbf{Minimum Description Length (MDL)}~\cite{DBLP:conf/emnlp/VoitaT20}: It quantifies the amount of effort required by a classifier to achieve certain accuracy. A higher value of MDL in our setting signifies that it is more difficult to retrieve the sensitive attribute from the encoded representation.  
% \end{itemize}
We provide more details about these metrics in Sec.~\ref{sec:appendix-privacy-metric}.

\begin{table*}[t]
    
    \centering
    % \adjustbox{max width=\linewidth}
    \begin{subtable}{\linewidth}
    \centering
        \begin{tabular}{lrrrr}
            \toprule
             Method                &     Accuracy $\uparrow$ &   TPR-gap $\downarrow$  &   Leakage $\downarrow$  &   MDL $\uparrow$\\
            \midrule
            \texttt{Random}        & 50.00 $\pm$ 0.00        &   0.00 $\pm$ 0.00       &   -                     &   31.3 $\pm$ 0.10\\
            \Uncon & 72.09 $\pm$ 0.73 & 26.26 $\pm$ 0.87  & 86.56 $\pm$ 0.83 & 15.21 $\pm$ 0.88\\
            \midrule \midrule
            \texttt{INLP} & 67.62 $\pm$ 0.57 &   9.19 $\pm$ 1.08  &   80.27 $\pm$ 2.50 & 24.82 $\pm$ 3.28 \\
            \texttt{Noise} & 71.52 $\pm$ 0.51 & 21.23 $\pm$ 2.50  & 66.29 $\pm$ 3.55 & 21.10 $\pm$ 1.81  \\
            \Adv         & 75.16 $\pm$ 0.65 & 5.03 $\pm$ 2.94  & 88.06 $\pm$ 0.20  & 16.16 $\pm$ 1.05  \\
             \AdvMultiple         & 75.32 $\pm$ 0.60  & 2.09 $\pm$ 1.18 & 88.03 $\pm$ 0.47 & 15.85 $\pm$ 1.46  \\
             
             \midrule \midrule
             
             \AdvNoise & 75.15 $\pm$ 0.59 & 1.75 $\pm$ 1.41  & 61.74 $\pm$ 5.05  & 22.94 $\pm$ 1.25 \\
            %  Adversarial + Differentiated + Noise & 74.61 &      1.79 &     17.44 &     62.1  &           54.42 \\
            \bottomrule
        \end{tabular}
        \caption{\label{tab:encoded-emoji-results-main} Results on Twitter Sentiment dataset. }
        % \label{tab:encoded-emoji-results-main}
    \end{subtable}\par

    \begin{subtable}{\linewidth}
    % \vspace*{0.5 cm}
        \centering
        \begin{tabular}{lrrrr}
        
            \toprule
             Method              &   Accuracy $\uparrow$ &     GRMS $\downarrow$ &  Leakage $\downarrow$   &  MDL $\uparrow$\\
            \midrule
            \texttt{Random} & 3.53 $\pm$ 0.01 &   0.00 $\pm$ 0.00 &   --   &   265.44 $\pm$ 0.13 \\
             \Uncon            & 79.29 $\pm$ 0.32 & 15.88 $\pm$ 0.80  & 75.92 $\pm$ 2.73 & 173.99 $\pm$ 7.08\\
             \midrule \midrule
             \texttt{INLP}         &     75.96  $\pm$ 0.47 & 12.81 $\pm$ 0.09 &  59.91 $\pm$ 0.08 & 253.36 $\pm$ 1.05\\
             \texttt{Noise}               & 77.88 $\pm$ 0.32 & 13.89 $\pm$ 0.31 & 62.23 $\pm$ 0.99 & 241.22 $\pm$ 2.97\\
             \Adv         & 79.02 $\pm$ 0.20  & 13.06 $\pm$ 0.39 & 69.47 $\pm$ 1.64 & 206.78 $\pm$ 13.02\\
             \AdvMultiple         & 79.30 $\pm$ 0.20  & 13.38 $\pm$ 0.63 & 68.24 $\pm$ 1.12 & 222.35 $\pm$ 10.04\\
             \midrule \midrule
             \AdvNoise & 77.79 $\pm$ 0.11 & 11.02 $\pm$ 0.55 & 56.92 $\pm$ 0.98 & 257.94 $\pm$ 1.93 \\
            \bottomrule
        \end{tabular}
        \caption{\label{tab:bias-in-bios-results-main} Results on Bias in Bios dataset.}
        % \label{tab:bias-in-bios-results-main}
    \end{subtable}
    \caption{Test results on (a) \textit{Twitter Sentiment}, and (b) \textit{Bias in Bios} with fixed Relaxation Threshold of 1.0. Fairness is measured with TPR-Gap or GRMS (lower is better), while privacy is measured by Leakage (lower is better) and MDL (higher is better). The MDL achieved by \texttt{Random} gives an upper bound for that particular dataset. Results have been averaged over $5$ different seeds. Our proposed \AdvNoise approach is the only method which achieves high levels of both fairness and privacy while maintaining competitive accuracy.}
\end{table*}

\paragraph{Methods and model architectures.} We compare  \textbf{\AdvNoise} to the following methods: (i) \textbf{\Adv} implements standard adversarial learning \cite{li-etal-2018-towards}, which is equivalent to our approach without the $priv$ layer, (ii) \textbf{\AdvMultiple} \cite{DBLP:conf/eacl/HanBC21} implements multiple adversaries, (iii) \textbf{\texttt{INLP}}~\cite{DBLP:conf/acl/RavfogelEGTG20} is a subspace projection approach, and (iv)
% GRAD-Pred~\cite{raff2018gradient}, where an adversary with an aim to learn sensitive attributes is added to an unconstrained classifier via a gradient reversal layer,
\textbf{\texttt{Noise}} learns DP text representations as proposed by \citet{DBLP:conf/emnlp/LyuHL20} but with corrected privacy analysis: this corresponds to our approach without the adversarial component. These methods have been described in details in Section~\ref{sec:related-work} and their hyperparametrs in Appendix~\ref{sec:hyper}.
We also report the performance of two simple baselines: \textbf{\texttt{Random}} simply predicts a random label, and \textbf{\Uncon} optimizes the classification performance without special consideration for privacy or fairness. 

% Note that \texttt{Noise} method is the same as~\citet{{DBLP:conf/emnlp/LyuHL20}} proposed but with correct L1-sensitivity. See Section~\ref{sec:related-work} for descriptions of \texttt{INLP} and \AdvMultiple.
% \textcolor{red}{We also do not evaluate \AdvMultiple over Bias in Bios as it is expensive to train and optimize due to presence of multiple adversaries.}

To provide a fair comparison, all methods use the same architecture for the encoder, the classifier and (when applicable) the adversarial branches.
In order to evaluate across varying model complexities, we employ different architectures for the different datasets.  For Twitter Sentiment, we follow the architecture employed by~\citet{DBLP:conf/eacl/HanBC21}, while for Bias in Bios we use a deeper architecture.
% which consists of a two-layer fully connected trainable encoder and a one-layer main task classifier. The adversarial branch consists of three-layer MLP.
The exact architecture, hyperparameters, and their tuning details are provided in Appendix~\ref{sec:architecture}-\ref{sec:hyper}. We implement \AdvNoise in PyTorch~\cite{NEURIPS2019_9015}. Our implementation, training, and evaluation scripts are available here.\footnote{The work-in-progress version of the codebase is currently available at \url{https://github.com/saist1993/DPNLP.}}

\subsection{Accuracy-Fairness-Privacy Trade-off}
\label{sec:relaxation}

% \gm{In this work, we are interested in approaches that are not only accurate but also fair and private at the same time. However, these three axes of evaluations, namely, privacy, fairness, and accuracy, are not independent and there exists an inherent tension between them. In this experiment, we explore this tridimensional trade-off which is influenced by the choice of method but also some of its hyperparameters (e.g., the value of $\epsilon$ and $\lambda$ in our approach).}
% As showcased by previous studies~\cite{DBLP:conf/fat/PujolMKHMM20,DBLP:conf/um/CummingsGKM19}, there exists an inherent tension between fairness, privacy, and accuracy. We also observed a similar tension between these axes through our experiments, and thus find it necessary to explore this tradeoff. In fact, we observed that hyperparameters that result in the highest validation accuracy, which is generally used as stopping criteria, does not always lead to a more fair model. Instead, hyperparameter around this validation accuracy often contains a more fair model, albeit slightly worse accuracy. 

In this first set of experiments, we explore the tridimensional trade-off between accuracy, fairness, and privacy and the inherent tension between them. These metrics are potentially all equally important and represent different information on different scales. Thus, they cannot be trivially combined into a single metric. Moreover, this trade-off is influenced by the choice of method but also some of its hyperparameters (e.g., the value of $\epsilon$ and $\lambda$ in our approach).
Previous studies \cite{DBLP:conf/eacl/HanBC21,DBLP:conf/emnlp/LyuHL20} essentially selected hyperparameter values that maximize validation accuracy, which may lead to undesirable or suboptimal trade-offs. For instance, we found that this strategy does not always induce a fairer model than the \Uncon baseline, and that it is often possible to obtain significantly more fair models at a negligible cost in accuracy.
% Previous studies~\cite{DBLP:conf/fat/PujolMKHMM20,DBLP:conf/um/CummingsGKM19} have shown an inherent tension between accuracy, privacy, and fairness, often at odds with another. In our experiments, we also observed a similar tension. For instance, we found that hyperparameters that result in the highest validation accuracy, commonly used as stopping criteria, do not always lead to a more fair model. Instead, hyperparameters around this validation accuracy often contain a more fair model, albeit with a slightly worse accuracy. On the other hand, choosing the hyperparameters with the lowest validation fairness scores and then exploring the region around them results in significantly worse accuracy across all approaches. Further, the drop in accuracy observed in the latter strategy (choosing lowest validation fairness scores) is more significant than the improvement in fairness scores observed in the former strategy.
Based on these observations, we propose to use a Relaxation Threshold (RT): %, which enables exploring these tradeoffs. Specifically,
instead of selecting the hyperparameters with highest validation accuracy $\alpha^*$, we consider all models with accuracy in the range $[\alpha^*-\text{RT}, \alpha^*]$. We then select the hyperparameters with best fairness score within that range.\footnote{
% This method can trivially be extended to include privacy as another factor.
% We can also incorporate privacy into our hyperparameter selection strategy but, for the datasets and methods in our study, we found no significant change in Leakage across the considered RT values, see Figure~\ref{fig:rt-tradeoff}.} 
We can also incorporate privacy into our hyperparameter selection strategy but, for the datasets and methods in our study, we found no significant change in Leakage across different hyperparameters.
% Thus we choose them based on accuracy and fairness.
}

Figure~\ref{fig:rt-tradeoff} presents the (validation) accuracy, fairness and privacy scores related to different RT for each method on Twitter Sentiment. The first thing to note is that \AdvNoise achieves the best fairness and privacy results with accuracy higher or comparable to competing approaches. We also observe that setting RT$=0.0$ (i.e., choosing the model with highest validation accuracy) leads to a significantly more unfair model in all approaches, while fairness generally improves with increasing RT. This improvement comes at a negligible or small cost in accuracy.
%We also find that for all adversarial approaches, the fairness gap amongst them decreases as we increase the RT.
In terms of privacy, we find no significant differences across RTs.

We now showcase detailed results with RT fixed to 1.0 which is found to provide good trade-offs for all approaches in Figure~\ref{fig:rt-tradeoff}, see Table~\ref{tab:encoded-emoji-results-main} for Twitter Sentiment and Table~\ref{tab:bias-in-bios-results-main} for Bias in Bios (and Appendix~\ref{sec:appendix-results} for additional results). %Detailed results on Twitter Sentiment dataset
For both datasets, we observe that all adversarial approaches induce a fairer model than \Uncon or \texttt{Noise}, with \AdvNoise performing best. In terms of accuracy, all adversarial approaches perform similarly on Twitter Sentiment. Interestingly, they achieve higher accuracy than \Uncon. We attribute this to a significant mismatch in the train and test distribution due to class imbalance. On Bias in Bios, we observe a small drop in accuracy of our proposed approach in comparison to \Adv, albeit with a corresponding gain in fairness. We hypothesize that this is due to the choice of possible hyperparameters for \AdvNoise (we did not consider very large values of $\epsilon$ which would recover \Adv), meaning that \AdvNoise pushes for more fairness (and privacy) at a potential cost of some accuracy. We explore the pairwise trade-offs (fairness-accuracy and privacy-accuracy) in more details in Section~\ref{sec:pairwise_tradeoffs}.

In terms of both privacy metrics,  \AdvNoise significantly outperforms all adversarial methods on both datasets. In fact, in line with previous studies~\cite{DBLP:conf/eacl/HanBC21}, the leakage and MDL of purely adversarial methods are similar to that of \Uncon.
% Over Bias in Bios, \texttt{INLP} provides a similar level of privacy to \AdvNoise, albeit with a worse accuracy. 
% \gmrm{However, in the case of Twitter Sentiment, our proposed approach leaks significantly less information while also having higher accuracy. }
On both datasets, \texttt{Noise} achieves slightly weaker privacy than \AdvNoise with much worse accuracy and fairness.

\AdvNoise also consistently outperforms \texttt{INLP} in all dimensions.

\textbf{In summary}, the results show that
% although some methods can provide either privacy or fairness,
\AdvNoise stands out as the only approach that can simultaneously induce a fairer model \emph{and} make its representation private while maintaining high accuracy. Furthermore, these results empirically demonstrate that our measures of privacy and fairness are indeed compatible with one another and can even reinforce each other.

% As showcased by previous studies, there exists an inherent tension between fairness, privacy and accuracy. Through our experiment we also observed a similar tension between these axes and found that hyperparameters that result in the highest validation accuracy do not always lead to a more fair model. Instead, hyperparameter around this validation accuracy often contains a more fair model, albeit with slightly smaller accuracy. Thus we propose, Relaxation Threshold (RT), which enables exploring these hyperparameter space. Specifically, instead of returning the model's hyperparameters with the highest validation accuracy ($\alpha^*$), we find all models with validation accuracy ($\alpha$) in the range $[\alpha^*-RF, \alpha^*]$. We then return the hyperparameters of the model with the best fairness scores within this range. 

\subsection{Pairwise Trade-offs}
\label{sec:pairwise_tradeoffs}

% \begin{figure}[t]
% \centering
% \includegraphics[width=\linewidth]{images/combined-acc-fairness-cropped-pdfcrop.pdf}
% \caption{Fairness-accuracy trade-off on Twitter Sentiment (top) and Bias in Bios (bottom).
% % The top graph corresponds to Bias in Bios and the bottom to Twitter Sentiment. \AdvNoise provides better fairness across most accuracy intervals in comparison to \Adv over both the datasets.
% A missing point means that the accuracy interval was not found within our hyperparameter search. }
% \label{fig:accuracy-fairness-graph}
% \end{figure}

\begin{figure}[t]
\centering
\includegraphics[width=.9\linewidth]{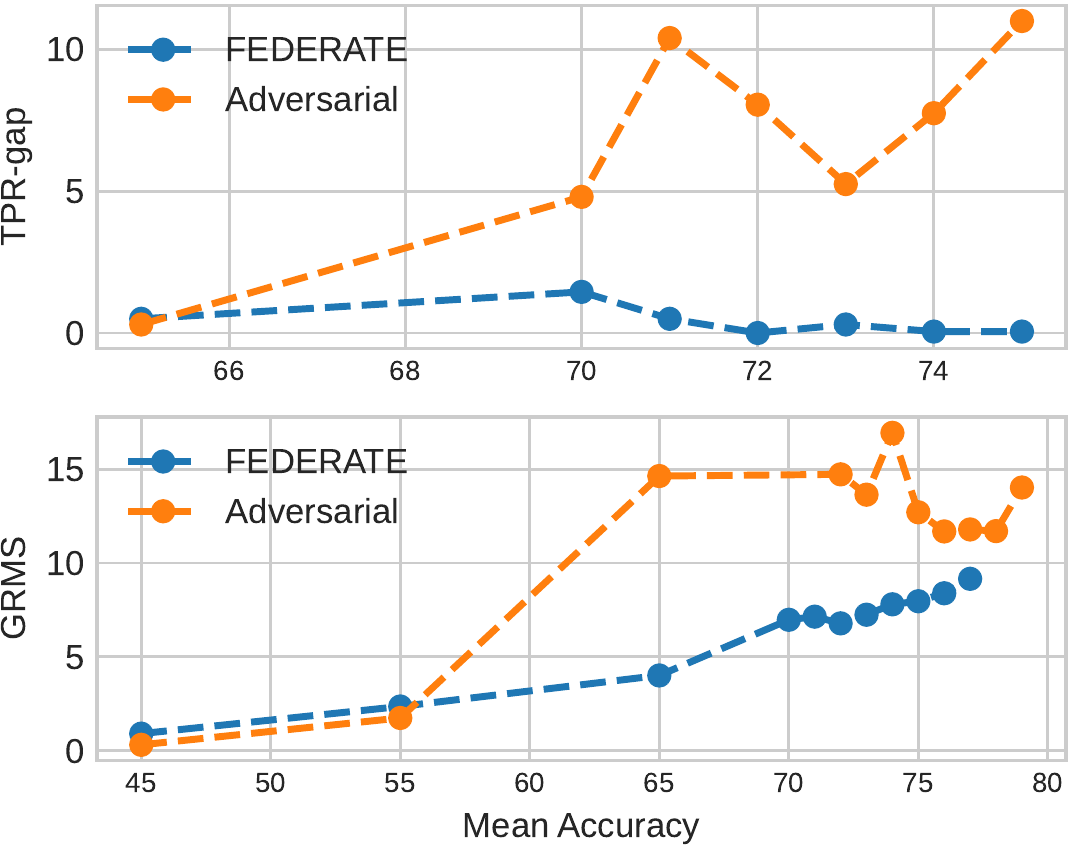}
\caption{Fairness-accuracy trade-off on Twitter Sentiment (top) and Bias in Bios (bottom).
% The top graph corresponds to Bias in Bios and the bottom to Twitter Sentiment.
A missing point means that the accuracy interval was not found within our hyperparameter search.
\AdvNoise provides better fairness across most accuracy intervals in comparison to \Adv over both datasets.}
\label{fig:accuracy-fairness-graph}
\end{figure}

% In the previous experiments, we considered accuracy, privacy, and fairness at the same time and found our approach to attain better trade-offs than all other approaches.
In the previous experiments, we explored the tridimensional trade-off and found~\AdvNoise to attain better trade-offs than all other methods.
Here, we take a closer look at the pairwise fairness-accuracy and privacy-accuracy trade-offs separately. We find that \AdvNoise outperforms the \Adv and \texttt{Noise} approach in their corresponding dimension, suggesting that \AdvNoise is a better choice even for bidimensional trade-offs. This experiment also validates the superiority of combining adversarial learning and DP over using either approach alone.

% make it more explicit in the para above that we are considering only two at a time. 
% \paragraph{Fairness-accuracy trade-off.} We plot best validation fairness scores over different accuracy intervals for the two datasets in Figure~\ref{fig:accuracy-fairness-graph}. The interval is denoted by mean accuracy, for instance, accuracy interval between $71.5$ and $72.5$ is represented with $72$. We then find the corresponding best fairness score for the interval. We find \AdvNoise provides better fairness than the \Adv approach for almost all the accuracy intervals. In the case of Bias in Bios, \Adv is able to achieve higher accuracy (albeit with a loss in fairness). We note that this high accuracy regime can be matched by \AdvNoise with a larger $\epsilon$.
% % We conjecture this to be the case because of the inherent tension between fairness and accuracy, where \AdvNoise focuses more on fairness. 
% % Increasing $\epsilon$ could cover this accuracy gap, albeit with a loss in fairness. 
% Interestingly, we find that \AdvNoise enables a smoother exploration of the accuracy-fairness trade-off space than \Adv, which shows much more erratic trajectories. Adversarial models are notoriously difficult to train, and this suggests that the introduction of DP noise has a stabilizing effect on the training dynamics of the adversarial component. 

\paragraph{Fairness-accuracy trade-off.} We plot best validation fairness scores over different accuracy intervals for the two datasets in Figure~\ref{fig:accuracy-fairness-graph}. The interval is denoted by its mean accuracy (i.e., $[71.5,72.5]$ is represented by $72$). We then find the corresponding best fairness score for the interval. We observe: 
\begin{itemize}
	\item \emph{Better fairness-accuracy trade-off:} \AdvNoise provides better fairness than the \Adv approach for almost all accuracy intervals. In the case of Bias in Bios, \Adv is able to achieve higher accuracy (albeit with a loss in fairness). We note that this high accuracy regime can be matched by \AdvNoise with a larger $\epsilon$.
	\item \emph{Smoother fairness-accuracy trade-off:} Interestingly, \AdvNoise enables a smoother exploration of the accuracy-fairness trade-off space than \Adv. %, which shows much more erratic trajectories.
	As adversarial models are notoriously difficult to train,  this suggests that the introduction of DP noise has a stabilizing effect on the training dynamics of the adversarial component. 
\end{itemize}

\begin{figure}[t]
\centering
\includegraphics[width=\linewidth]{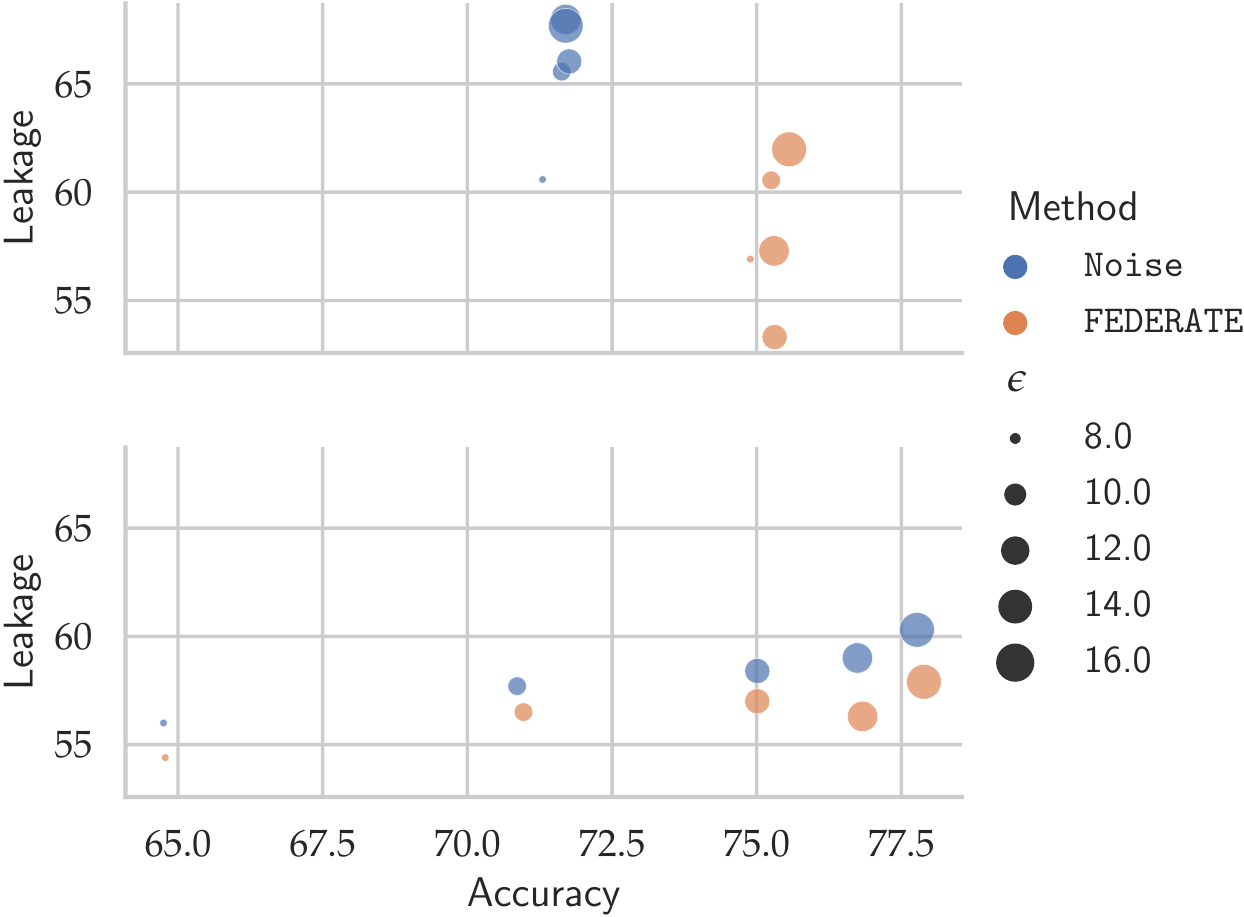}
\caption{Privacy-accuracy trade-off on Twitter Sentiment (top) and Bias in Bios (bottom), with associated values of $\epsilon$.
\AdvNoise gives lower leakage and better or comparable accuracy to \texttt{Noise} over both datasets.
}
\label{fig:accuracy-privacy-graph}
\end{figure}

\paragraph{Privacy-accuracy trade-off.} We plot privacy and accuracy with respect to $\epsilon$, the parameter controlling the theoretical privacy level in Figure~\ref{fig:accuracy-privacy-graph}. In general, the value of $\epsilon$ correlates well with the empirical leakage. On Bias in Bios, \AdvNoise and \texttt{Noise} are comparable in both accuracy and privacy. However, for Twitter Sentiment, our approach outperforms \texttt{Noise} in both accuracy and privacy for every $\epsilon$. We hypothesize this difference in the accuracy to be a case of mismatch between train-test split, suggesting \AdvNoise to be more robust to these distributional shifts. These observations suggest that \AdvNoise either improves upon \texttt{Noise} in privacy-accuracy tradeoff or remains comparable. For completeness, we also present the same results as a table in Appendix~\ref{sec:appendix-results}.

\section{Conclusion and Perspectives}
\label{sec:conclu}

We proposed a DP-driven adversarial learning approach for NLP. Through our experiments, we showed that our method simultaneously induces private representations and fair models, with a mutually reinforcing effect between privacy and fairness. We also find that our approach improves upon competitors on each dimension separately. While we focused on privatizing sensitive attributes like race or gender, our approach can be used to remove other types of unwanted information from text representations, such as tenses or POS tag information, which might not be relevant for certain NLP tasks.  

% A current limitation of this work in context of fairness is that it is not designed to work with a specific definition of fairness, such as equal odds.
A possible limitation of this work is that it not tailored to a specific definition of fairness like equal odds. Instead, it enforces fairness by removing certain protected information, which can correlate with specific fairness notions. Similarly, we do not provide any formal fairness guarantees for our method, as we do for privacy. In the future, we aim to investigate fairness methods that explicitly optimize for a specific fairness definition and explore other privacy threats (e.g., reconstruction attacks).

\section{Acknowledgement}
The authors would like to thank the Agence Nationale de la Recherche for funding this work under grant number ANR-19-CE23-0022, as well as the ARR reviewers for their feedback and suggestions.

\bibliographystyle{acl_natbib}
\bibliography{anthology,acl2021}

\begin{thebibliography}{51}
\expandafter\ifx\csname natexlab\endcsname\relax\def\natexlab#1{#1}\fi

\bibitem[{Abowd(2018)}]{abowd2018us}
John~M Abowd. 2018.
\newblock The us census bureau adopts differential privacy.
\newblock In \emph{Proceedings of the 24th ACM SIGKDD International Conference
  on Knowledge Discovery \& Data Mining}, pages 2867--2867.

\bibitem[{Adi et~al.(2019)Adi, Zeghidour, Collobert, Usunier, Liptchinsky, and
  Synnaeve}]{DBLP:conf/icassp/AdiZCULS19}
Yossi Adi, Neil Zeghidour, Ronan Collobert, Nicolas Usunier, Vitaliy
  Liptchinsky, and Gabriel Synnaeve. 2019.
\newblock \href {https://doi.org/10.1109/ICASSP.2019.8682468} {To reverse the
  gradient or not: an empirical comparison of adversarial and multi-task
  learning in speech recognition}.
\newblock In \emph{{IEEE} International Conference on Acoustics, Speech and
  Signal Processing, {ICASSP} 2019, Brighton, United Kingdom, May 12-17, 2019},
  pages 3742--3746. {IEEE}.

\bibitem[{Bagdasaryan et~al.(2019)Bagdasaryan, Poursaeed, and
  Shmatikov}]{DBLP:conf/nips/BagdasaryanPS19}
Eugene Bagdasaryan, Omid Poursaeed, and Vitaly Shmatikov. 2019.
\newblock \href
  {https://proceedings.neurips.cc/paper/2019/hash/fc0de4e0396fff257ea362983c2dda5a-Abstract.html}
  {Differential privacy has disparate impact on model accuracy}.
\newblock In \emph{Advances in Neural Information Processing Systems 32: Annual
  Conference on Neural Information Processing Systems 2019, NeurIPS 2019,
  December 8-14, 2019, Vancouver, BC, Canada}, pages 15453--15462.

\bibitem[{Blodgett et~al.(2016)Blodgett, Green, and
  O{'}Connor}]{blodgett-etal-2016-demographic}
Su~Lin Blodgett, Lisa Green, and Brendan O{'}Connor. 2016.
\newblock \href {https://doi.org/10.18653/v1/D16-1120} {Demographic dialectal
  variation in social media: A case study of {A}frican-{A}merican {E}nglish}.
\newblock In \emph{Proceedings of the 2016 Conference on Empirical Methods in
  Natural Language Processing}, pages 1119--1130, Austin, Texas. Association
  for Computational Linguistics.

\bibitem[{Bolukbasi et~al.(2016)Bolukbasi, Chang, Zou, Saligrama, and
  Kalai}]{DBLP:conf/nips/BolukbasiCZSK16}
Tolga Bolukbasi, Kai{-}Wei Chang, James~Y. Zou, Venkatesh Saligrama, and
  Adam~Tauman Kalai. 2016.
\newblock \href
  {https://proceedings.neurips.cc/paper/2016/hash/a486cd07e4ac3d270571622f4f316ec5-Abstract.html}
  {Man is to computer programmer as woman is to homemaker? debiasing word
  embeddings}.
\newblock In \emph{Advances in Neural Information Processing Systems 29: Annual
  Conference on Neural Information Processing Systems 2016, December 5-10,
  2016, Barcelona, Spain}, pages 4349--4357.

\bibitem[{Chang and Shokri(2020)}]{DBLP:journals/corr/abs-2011-03731}
Hongyan Chang and Reza Shokri. 2020.
\newblock \href {http://arxiv.org/abs/2011.03731} {On the privacy risks of
  algorithmic fairness}.
\newblock \emph{CoRR}, abs/2011.03731.

\bibitem[{Chowdhury et~al.(2021)Chowdhury, Ghosh, Li, Oliva, Srivastava, and
  Chaturvedi}]{DBLP:conf/emnlp/ChowdhuryGLOSC21}
Somnath Basu~Roy Chowdhury, Sayan Ghosh, Yiyuan Li, Junier Oliva, Shashank
  Srivastava, and Snigdha Chaturvedi. 2021.
\newblock \href {https://doi.org/10.18653/v1/2021.emnlp-main.43} {Adversarial
  scrubbing of demographic information for text classification}.
\newblock In \emph{Proceedings of the 2021 Conference on Empirical Methods in
  Natural Language Processing, {EMNLP} 2021, Virtual Event / Punta Cana,
  Dominican Republic, 7-11 November, 2021}, pages 550--562. Association for
  Computational Linguistics.

\bibitem[{Coavoux et~al.(2018)Coavoux, Narayan, and
  Cohen}]{DBLP:conf/emnlp/CoavouxNC18}
Maximin Coavoux, Shashi Narayan, and Shay~B. Cohen. 2018.
\newblock \href {https://doi.org/10.18653/v1/d18-1001} {Privacy-preserving
  neural representations of text}.
\newblock In \emph{Proceedings of the 2018 Conference on Empirical Methods in
  Natural Language Processing, Brussels, Belgium, October 31 - November 4,
  2018}, pages 1--10. Association for Computational Linguistics.

\bibitem[{Cummings et~al.(2019)Cummings, Gupta, Kimpara, and
  Morgenstern}]{DBLP:conf/um/CummingsGKM19}
Rachel Cummings, Varun Gupta, Dhamma Kimpara, and Jamie Morgenstern. 2019.
\newblock \href {https://doi.org/10.1145/3314183.3323847} {On the compatibility
  of privacy and fairness}.
\newblock In \emph{Adjunct Publication of the 27th Conference on User Modeling,
  Adaptation and Personalization, {UMAP} 2019, Larnaca, Cyprus, June 09-12,
  2019}, pages 309--315. {ACM}.

\bibitem[{De{-}Arteaga et~al.(2019)De{-}Arteaga, Romanov, Wallach, Chayes,
  Borgs, Chouldechova, Geyik, Kenthapadi, and
  Kalai}]{DBLP:conf/fat/De-ArteagaRWCBC19}
Maria De{-}Arteaga, Alexey Romanov, Hanna~M. Wallach, Jennifer~T. Chayes,
  Christian Borgs, Alexandra Chouldechova, Sahin~Cem Geyik, Krishnaram
  Kenthapadi, and Adam~Tauman Kalai. 2019.
\newblock \href {https://doi.org/10.1145/3287560.3287572} {Bias in bios: {A}
  case study of semantic representation bias in a high-stakes setting}.
\newblock In \emph{Proceedings of the Conference on Fairness, Accountability,
  and Transparency, FAT* 2019, Atlanta, GA, USA, January 29-31, 2019}, pages
  120--128. {ACM}.

\bibitem[{Devlin et~al.(2019)Devlin, Chang, Lee, and
  Toutanova}]{DBLP:conf/naacl/DevlinCLT19}
Jacob Devlin, Ming{-}Wei Chang, Kenton Lee, and Kristina Toutanova. 2019.
\newblock \href {https://doi.org/10.18653/v1/n19-1423} {{BERT:} pre-training of
  deep bidirectional transformers for language understanding}.
\newblock In \emph{Proceedings of the 2019 Conference of the North American
  Chapter of the Association for Computational Linguistics: Human Language
  Technologies, {NAACL-HLT} 2019, Minneapolis, MN, USA, June 2-7, 2019, Volume
  1 (Long and Short Papers)}, pages 4171--4186. Association for Computational
  Linguistics.

\bibitem[{Ding et~al.(2017)Ding, Kulkarni, and Yekhanin}]{telemetry:17}
Bolin Ding, Janardhan Kulkarni, and Sergey Yekhanin. 2017.
\newblock Collecting telemetry data privately.
\newblock In \emph{NIPS}.

\bibitem[{Dua and Graff(2017)}]{Dua:2019}
Dheeru Dua and Casey Graff. 2017.
\newblock \href {http://archive.ics.uci.edu/ml} {{UCI} machine learning
  repository}.

\bibitem[{Dwork et~al.(2006)Dwork, McSherry, Nissim, and
  Smith}]{DBLP:conf/tcc/DworkMNS06}
Cynthia Dwork, Frank McSherry, Kobbi Nissim, and Adam~D. Smith. 2006.
\newblock \href {https://doi.org/10.1007/11681878\_14} {Calibrating noise to
  sensitivity in private data analysis}.
\newblock In \emph{Theory of Cryptography, Third Theory of Cryptography
  Conference, {TCC} 2006, New York, NY, USA, March 4-7, 2006, Proceedings},
  volume 3876 of \emph{Lecture Notes in Computer Science}, pages 265--284.
  Springer.

\bibitem[{Dwork and Roth(2014)}]{DBLP:journals/fttcs/DworkR14}
Cynthia Dwork and Aaron Roth. 2014.
\newblock \href {https://doi.org/10.1561/0400000042} {The algorithmic
  foundations of differential privacy}.
\newblock \emph{Found. Trends Theor. Comput. Sci.}, 9(3-4):211--407.

\bibitem[{Elazar and Goldberg(2018)}]{DBLP:conf/emnlp/ElazarG18}
Yanai Elazar and Yoav Goldberg. 2018.
\newblock \href {https://doi.org/10.18653/v1/d18-1002} {Adversarial removal of
  demographic attributes from text data}.
\newblock In \emph{Proceedings of the 2018 Conference on Empirical Methods in
  Natural Language Processing, Brussels, Belgium, October 31 - November 4,
  2018}, pages 11--21. Association for Computational Linguistics.

\bibitem[{Erlingsson et~al.(2014)Erlingsson, Pihur, and Korolova}]{rappor:15}
\'{U}lfar Erlingsson, Vasyl Pihur, and Aleksandra Korolova. 2014.
\newblock Rappor: Randomized aggregatable privacy-preserving ordinal response.
\newblock In \emph{{CCS}}.

\bibitem[{Fanti et~al.(2016)Fanti, Pihur, and Erlingsson}]{rappor2:16}
Giulia Fanti, Vasyl Pihur, and {\'U}lfar Erlingsson. 2016.
\newblock Building a {RAPPOR} with the unknown: {P}rivacy-preserving learning
  of associations and data dictionaries.
\newblock In \emph{PoPETs}.

\bibitem[{Felbo et~al.(2017)Felbo, Mislove, S{\o}gaard, Rahwan, and
  Lehmann}]{DBLP:conf/emnlp/FelboMSRL17}
Bjarke Felbo, Alan Mislove, Anders S{\o}gaard, Iyad Rahwan, and Sune Lehmann.
  2017.
\newblock \href {https://doi.org/10.18653/v1/d17-1169} {Using millions of emoji
  occurrences to learn any-domain representations for detecting sentiment,
  emotion and sarcasm}.
\newblock In \emph{Proceedings of the 2017 Conference on Empirical Methods in
  Natural Language Processing, {EMNLP} 2017, Copenhagen, Denmark, September
  9-11, 2017}, pages 1615--1625. Association for Computational Linguistics.

\bibitem[{Ganin and Lempitsky(2015)}]{pmlr-v37-ganin15}
Yaroslav Ganin and Victor Lempitsky. 2015.
\newblock \href {https://proceedings.mlr.press/v37/ganin15.html} {Unsupervised
  domain adaptation by backpropagation}.
\newblock In \emph{Proceedings of the 32nd International Conference on Machine
  Learning}, volume~37 of \emph{Proceedings of Machine Learning Research},
  pages 1180--1189, Lille, France. PMLR.

\bibitem[{Gonen and Goldberg(2019)}]{gonen-goldberg-2019-lipstick}
Hila Gonen and Yoav Goldberg. 2019.
\newblock \href {https://www.aclweb.org/anthology/W19-3621} {Lipstick on a pig:
  Debiasing methods cover up systematic gender biases in word embeddings but do
  not remove them}.
\newblock In \emph{Proceedings of the 2019 Workshop on Widening NLP}, pages
  60--63, Florence, Italy. Association for Computational Linguistics.

\bibitem[{Habernal(2021)}]{DBLP:conf/emnlp/Habernal21}
Ivan Habernal. 2021.
\newblock \href {https://doi.org/10.18653/v1/2021.emnlp-main.114} {When
  differential privacy meets {NLP:} the devil is in the detail}.
\newblock In \emph{Proceedings of the 2021 Conference on Empirical Methods in
  Natural Language Processing, {EMNLP} 2021, Virtual Event / Punta Cana,
  Dominican Republic, 7-11 November, 2021}, pages 1522--1528. Association for
  Computational Linguistics.

\bibitem[{Han et~al.(2021)Han, Baldwin, and Cohn}]{DBLP:conf/eacl/HanBC21}
Xudong Han, Timothy Baldwin, and Trevor Cohn. 2021.
\newblock \href {https://aclanthology.org/2021.eacl-main.239/} {Diverse
  adversaries for mitigating bias in training}.
\newblock In \emph{Proceedings of the 16th Conference of the European Chapter
  of the Association for Computational Linguistics: Main Volume, {EACL} 2021,
  Online, April 19 - 23, 2021}, pages 2760--2765. Association for Computational
  Linguistics.

\bibitem[{Jagielski et~al.(2019)Jagielski, Kearns, Mao, Oprea, Roth,
  Sharifi{-}Malvajerdi, and Ullman}]{DBLP:conf/icml/JagielskiKMORSU19}
Matthew Jagielski, Michael~J. Kearns, Jieming Mao, Alina Oprea, Aaron Roth,
  Saeed Sharifi{-}Malvajerdi, and Jonathan~R. Ullman. 2019.
\newblock \href {http://proceedings.mlr.press/v97/jagielski19a.html}
  {Differentially private fair learning}.
\newblock In \emph{Proceedings of the 36th International Conference on Machine
  Learning, {ICML} 2019, 9-15 June 2019, Long Beach, California, {USA}},
  volume~97 of \emph{Proceedings of Machine Learning Research}, pages
  3000--3008. {PMLR}.

\bibitem[{Karve et~al.(2019)Karve, Ungar, and
  Sedoc}]{karve-etal-2019-conceptor}
Saket Karve, Lyle Ungar, and Jo{\~a}o Sedoc. 2019.
\newblock \href {https://doi.org/10.18653/v1/W19-3806} {Conceptor debiasing of
  word representations evaluated on {WEAT}}.
\newblock In \emph{Proceedings of the First Workshop on Gender Bias in Natural
  Language Processing}, pages 40--48, Florence, Italy. Association for
  Computational Linguistics.

\bibitem[{Kasiviswanathan et~al.(2011)Kasiviswanathan, Lee, Nissim,
  Raskhodnikova, and Smith}]{DBLP:journals/siamcomp/KasiviswanathanLNRS11}
Shiva~Prasad Kasiviswanathan, Homin~K. Lee, Kobbi Nissim, Sofya Raskhodnikova,
  and Adam~D. Smith. 2011.
\newblock \href {https://doi.org/10.1137/090756090} {What can we learn
  privately?}
\newblock \emph{{SIAM} J. Comput.}, 40(3):793--826.

\bibitem[{Kiritchenko and Mohammad(2018)}]{kiritchenko-mohammad-2018-examining}
Svetlana Kiritchenko and Saif Mohammad. 2018.
\newblock \href {https://doi.org/10.18653/v1/S18-2005} {Examining gender and
  race bias in two hundred sentiment analysis systems}.
\newblock In \emph{Proceedings of the Seventh Joint Conference on Lexical and
  Computational Semantics}, pages 43--53, New Orleans, Louisiana. Association
  for Computational Linguistics.

\bibitem[{Kohavi(1996)}]{DBLP:conf/kdd/Kohavi96}
Ron Kohavi. 1996.
\newblock \href {http://www.aaai.org/Library/KDD/1996/kdd96-033.php} {Scaling
  up the accuracy of naive-bayes classifiers: {A} decision-tree hybrid}.
\newblock In \emph{Proceedings of the Second International Conference on
  Knowledge Discovery and Data Mining (KDD-96), Portland, Oregon, {USA}}, pages
  202--207. {AAAI} Press.

\bibitem[{Krishna et~al.(2021)Krishna, Gupta, and
  Dupuy}]{krishna-etal-2021-adept}
Satyapriya Krishna, Rahul Gupta, and Christophe Dupuy. 2021.
\newblock \href {https://doi.org/10.18653/v1/2021.eacl-main.207} {{AD}e{PT}:
  Auto-encoder based differentially private text transformation}.
\newblock In \emph{Proceedings of the 16th Conference of the European Chapter
  of the Association for Computational Linguistics: Main Volume}, pages
  2435--2439, Online. Association for Computational Linguistics.

\bibitem[{Kulynych et~al.(2022)Kulynych, Yaghini, Cherubin, Veale, and
  Troncoso}]{DBLP:journals/corr/abs-1906-00389}
Bogdan Kulynych, Mohammad Yaghini, Giovanni Cherubin, Michael Veale, and
  Carmela Troncoso. 2022.
\newblock \href {http://arxiv.org/abs/1906.00389} {Disparate vulnerability to
  membership inference attacks}.
\newblock In \emph{PETS}.

\bibitem[{Lample et~al.(2017)Lample, Zeghidour, Usunier, Bordes, Denoyer, and
  Ranzato}]{DBLP:conf/nips/LampleZUBDR17}
Guillaume Lample, Neil Zeghidour, Nicolas Usunier, Antoine Bordes, Ludovic
  Denoyer, and Marc'Aurelio Ranzato. 2017.
\newblock \href
  {https://proceedings.neurips.cc/paper/2017/hash/3fd60983292458bf7dee75f12d5e9e05-Abstract.html}
  {Fader networks: Manipulating images by sliding attributes}.
\newblock In \emph{Advances in Neural Information Processing Systems 30: Annual
  Conference on Neural Information Processing Systems 2017, December 4-9, 2017,
  Long Beach, CA, {USA}}, pages 5967--5976.

\bibitem[{Li et~al.(2018)Li, Baldwin, and Cohn}]{li-etal-2018-towards}
Yitong Li, Timothy Baldwin, and Trevor Cohn. 2018.
\newblock \href {https://doi.org/10.18653/v1/P18-2005} {Towards robust and
  privacy-preserving text representations}.
\newblock In \emph{Proceedings of the 56th Annual Meeting of the Association
  for Computational Linguistics (Volume 2: Short Papers)}, pages 25--30,
  Melbourne, Australia. Association for Computational Linguistics.

\bibitem[{Liu et~al.(2020)Liu, Wang, Lu, Cheng, Jin, Wang, and
  Zha}]{liu2020fair}
Wenyan Liu, Xiangfeng Wang, Xingjian Lu, Junhong Cheng, Bo~Jin, Xiaoling Wang,
  and Hongyuan Zha. 2020.
\newblock Fair differential privacy can mitigate the disparate impact on model
  accuracy.

\bibitem[{Liu et~al.(2015)Liu, Luo, Wang, and Tang}]{DBLP:conf/iccv/LiuLWT15}
Ziwei Liu, Ping Luo, Xiaogang Wang, and Xiaoou Tang. 2015.
\newblock \href {https://doi.org/10.1109/ICCV.2015.425} {Deep learning face
  attributes in the wild}.
\newblock In \emph{2015 {IEEE} International Conference on Computer Vision,
  {ICCV} 2015, Santiago, Chile, December 7-13, 2015}, pages 3730--3738. {IEEE}
  Computer Society.

\bibitem[{Lohaus et~al.(2020)Lohaus, Perrot, and Luxburg}]{pmlr-v119-lohaus20a}
Michael Lohaus, Michael Perrot, and Ulrike~Von Luxburg. 2020.
\newblock \href {https://proceedings.mlr.press/v119/lohaus20a.html} {Too
  relaxed to be fair}.
\newblock In \emph{Proceedings of the 37th International Conference on Machine
  Learning}, volume 119 of \emph{Proceedings of Machine Learning Research},
  pages 6360--6369. PMLR.

\bibitem[{Lyu et~al.(2020)Lyu, He, and Li}]{DBLP:conf/emnlp/LyuHL20}
Lingjuan Lyu, Xuanli He, and Yitong Li. 2020.
\newblock \href {https://doi.org/10.18653/v1/2020.findings-emnlp.213}
  {Differentially private representation for {NLP:} formal guarantee and an
  empirical study on privacy and fairness}.
\newblock In \emph{Findings of the Association for Computational Linguistics:
  {EMNLP} 2020, Online Event, 16-20 November 2020}, volume {EMNLP} 2020 of
  \emph{Findings of {ACL}}, pages 2355--2365. Association for Computational
  Linguistics.

\bibitem[{Paszke et~al.(2019)Paszke, Gross, Massa, Lerer, Bradbury, Chanan,
  Killeen, Lin, Gimelshein, Antiga, Desmaison, Kopf, Yang, DeVito, Raison,
  Tejani, Chilamkurthy, Steiner, Fang, Bai, and Chintala}]{NEURIPS2019_9015}
Adam Paszke, Sam Gross, Francisco Massa, Adam Lerer, James Bradbury, Gregory
  Chanan, Trevor Killeen, Zeming Lin, Natalia Gimelshein, Luca Antiga, Alban
  Desmaison, Andreas Kopf, Edward Yang, Zachary DeVito, Martin Raison, Alykhan
  Tejani, Sasank Chilamkurthy, Benoit Steiner, Lu~Fang, Junjie Bai, and Soumith
  Chintala. 2019.
\newblock \href
  {http://papers.neurips.cc/paper/9015-pytorch-an-imperative-style-high-performance-deep-learning-library.pdf}
  {Pytorch: An imperative style, high-performance deep learning library}.
\newblock In H.~Wallach, H.~Larochelle, A.~Beygelzimer, F.~d\textquotesingle
  Alch\'{e}-Buc, E.~Fox, and R.~Garnett, editors, \emph{Advances in Neural
  Information Processing Systems 32}, pages 8024--8035. Curran Associates, Inc.

\bibitem[{Plant et~al.(2021)Plant, Gkatzia, and Giuffrida}]{plant2021cape}
Richard Plant, Dimitra Gkatzia, and Valerio Giuffrida. 2021.
\newblock Cape: Context-aware private embeddings for private language learning.
\newblock \emph{arXiv preprint arXiv:2108.12318}.

\bibitem[{Pujol et~al.(2020)Pujol, McKenna, Kuppam, Hay, Machanavajjhala, and
  Miklau}]{DBLP:conf/fat/PujolMKHMM20}
David Pujol, Ryan McKenna, Satya Kuppam, Michael Hay, Ashwin Machanavajjhala,
  and Gerome Miklau. 2020.
\newblock \href {https://doi.org/10.1145/3351095.3372872} {Fair decision making
  using privacy-protected data}.
\newblock In \emph{FAT* '20: Conference on Fairness, Accountability, and
  Transparency, Barcelona, Spain, January 27-30, 2020}, pages 189--199. {ACM}.

\bibitem[{Raghavan et~al.(2020)Raghavan, Barocas, Kleinberg, and
  Levy}]{DBLP:conf/fat/RaghavanBKL20}
Manish Raghavan, Solon Barocas, Jon~M. Kleinberg, and Karen Levy. 2020.
\newblock \href {https://doi.org/10.1145/3351095.3372828} {Mitigating bias in
  algorithmic hiring: evaluating claims and practices}.
\newblock In \emph{FAT* '20: Conference on Fairness, Accountability, and
  Transparency, Barcelona, Spain, January 27-30, 2020}, pages 469--481. {ACM}.

\bibitem[{Ravfogel et~al.(2020)Ravfogel, Elazar, Gonen, Twiton, and
  Goldberg}]{DBLP:conf/acl/RavfogelEGTG20}
Shauli Ravfogel, Yanai Elazar, Hila Gonen, Michael Twiton, and Yoav Goldberg.
  2020.
\newblock \href {https://doi.org/10.18653/v1/2020.acl-main.647} {Null it out:
  Guarding protected attributes by iterative nullspace projection}.
\newblock In \emph{Proceedings of the 58th Annual Meeting of the Association
  for Computational Linguistics, {ACL} 2020, Online, July 5-10, 2020}, pages
  7237--7256. Association for Computational Linguistics.

\bibitem[{Romanov et~al.(2019)Romanov, De-Arteaga, Wallach, Chayes, Borgs,
  Chouldechova, Geyik, Kenthapadi, Rumshisky, and
  Kalai}]{romanov-etal-2019-whats}
Alexey Romanov, Maria De-Arteaga, Hanna Wallach, Jennifer Chayes, Christian
  Borgs, Alexandra Chouldechova, Sahin Geyik, Krishnaram Kenthapadi, Anna
  Rumshisky, and Adam Kalai. 2019.
\newblock \href {https://doi.org/10.18653/v1/N19-1424} {What{'}s in a name?
  {R}educing bias in bios without access to protected attributes}.
\newblock In \emph{Proceedings of the 2019 Conference of the North {A}merican
  Chapter of the Association for Computational Linguistics: Human Language
  Technologies, Volume 1 (Long and Short Papers)}, pages 4187--4195,
  Minneapolis, Minnesota. Association for Computational Linguistics.

\bibitem[{Shokri and Shmatikov(2015)}]{priv_deep_learning}
Reza Shokri and Vitaly Shmatikov. 2015.
\newblock Privacy-preserving deep learning.
\newblock In \emph{{CCS}}.

\bibitem[{Shokri et~al.(2017)Shokri, Stronati, Song, and
  Shmatikov}]{DBLP:conf/sp/ShokriSSS17}
Reza Shokri, Marco Stronati, Congzheng Song, and Vitaly Shmatikov. 2017.
\newblock \href {https://doi.org/10.1109/SP.2017.41} {Membership inference
  attacks against machine learning models}.
\newblock In \emph{2017 {IEEE} Symposium on Security and Privacy, {SP} 2017,
  San Jose, CA, USA, May 22-26, 2017}, pages 3--18. {IEEE} Computer Society.

\bibitem[{Song and Raghunathan(2020)}]{DBLP:conf/ccs/SongR20}
Congzheng Song and Ananth Raghunathan. 2020.
\newblock \href {https://doi.org/10.1145/3372297.3417270} {Information leakage
  in embedding models}.
\newblock In \emph{{CCS} '20: 2020 {ACM} {SIGSAC} Conference on Computer and
  Communications Security, Virtual Event, USA, November 9-13, 2020}, pages
  377--390. {ACM}.

\bibitem[{van~den Broek et~al.(2019)van~den Broek, Sergeeva, and
  Huysman}]{DBLP:conf/icis/BroekSH19}
Elmira van~den Broek, Anastasia Sergeeva, and Marleen Huysman. 2019.
\newblock \href
  {https://aisel.aisnet.org/icis2019/future\_of\_work/future\_work/6} {Hiring
  algorithms: An ethnography of fairness in practice}.
\newblock In \emph{Proceedings of the 40th International Conference on
  Information Systems, {ICIS} 2019, Munich, Germany, December 15-18, 2019}.
  Association for Information Systems.

\bibitem[{Voita and Titov(2020)}]{DBLP:conf/emnlp/VoitaT20}
Elena Voita and Ivan Titov. 2020.
\newblock \href {https://doi.org/10.18653/v1/2020.emnlp-main.14}
  {Information-theoretic probing with minimum description length}.
\newblock In \emph{Proceedings of the 2020 Conference on Empirical Methods in
  Natural Language Processing, {EMNLP} 2020, Online, November 16-20, 2020},
  pages 183--196. Association for Computational Linguistics.

\bibitem[{Wang et~al.(2020)Wang, Lin, Rajani, McCann, Ordonez, and
  Xiong}]{DBLP:conf/acl/WangLRMOX20}
Tianlu Wang, Xi~Victoria Lin, Nazneen~Fatema Rajani, Bryan McCann, Vicente
  Ordonez, and Caiming Xiong. 2020.
\newblock \href {https://doi.org/10.18653/v1/2020.acl-main.484} {Double-hard
  debias: Tailoring word embeddings for gender bias mitigation}.
\newblock In \emph{Proceedings of the 58th Annual Meeting of the Association
  for Computational Linguistics, {ACL} 2020, Online, July 5-10, 2020}, pages
  5443--5453. Association for Computational Linguistics.

\bibitem[{Wu et~al.(2019)Wu, Zhang, and Wu}]{DBLP:conf/www/Wu0W19}
Yongkai Wu, Lu~Zhang, and Xintao Wu. 2019.
\newblock \href {https://doi.org/10.1145/3308558.3313723} {On convexity and
  bounds of fairness-aware classification}.
\newblock In \emph{The World Wide Web Conference, {WWW} 2019, San Francisco,
  CA, USA, May 13-17, 2019}, pages 3356--3362. {ACM}.

\bibitem[{Xu et~al.(2020)Xu, Du, and Wu}]{DBLP:journals/corr/abs-2003-03699}
Depeng Xu, Wei Du, and Xintao Wu. 2020.
\newblock \href {http://arxiv.org/abs/2003.03699} {Removing disparate impact of
  differentially private stochastic gradient descent on model accuracy}.
\newblock \emph{CoRR}, abs/2003.03699.

\bibitem[{Zhao et~al.(2018)Zhao, Wang, Yatskar, Ordonez, and
  Chang}]{DBLP:conf/naacl/ZhaoWYOC18}
Jieyu Zhao, Tianlu Wang, Mark Yatskar, Vicente Ordonez, and Kai{-}Wei Chang.
  2018.
\newblock \href {https://doi.org/10.18653/v1/n18-2003} {Gender bias in
  coreference resolution: Evaluation and debiasing methods}.
\newblock In \emph{Proceedings of the 2018 Conference of the North American
  Chapter of the Association for Computational Linguistics: Human Language
  Technologies, NAACL-HLT, New Orleans, Louisiana, USA, June 1-6, 2018, Volume
  2 (Short Papers)}, pages 15--20. Association for Computational Linguistics.

\end{thebibliography}

\clearpage
\appendix

\section*{APPENDIX}

\section{Training Algorithm}
\label{sec:algorithm}

We provide the pseudo-code of the training procedure of \AdvNoise in Algorithm~\ref{alg:algo}.
Note that the combination of Steps 2-3-4 corresponds to $E_{priv}$ in Sec.~\ref{sec:approach}.

\begin{algorithm*}[t]
\DontPrintSemicolon
  
  \KwInput{ Model architecture composed of encoder $E$ (parameterized by $\theta_E$), classifier $C$ (parameterized by $\theta_C$), adversary $A$ (parameterized by $\theta_A$), loss function $L$
    }
  \KwOutput{Trained model}
  \KwData{Samples S=${\{x^i,y^i,z^i\}_{i=1}^{m}}$ where $x^i$ is the input text, $y^i$ is the task label, and $z^i$ is the sensitive attribute. }
    \For{$i\gets0$ \KwTo $m$}{ \tcp{For each sample in the dataset. This can be batch too. }
    Encode: $\textbf{x}^i \leftarrow E(x^i)$\\
    Normalize: $\textbf{x}^i \leftarrow \frac{\textbf{x}^i}{\|\textbf{x}^i\|_1}$\\
    Privatize: $\textbf{x}^i_{priv} \leftarrow \textbf{x}^i  +  \boldsymbol{\ell}$, where each entry of the vector $\boldsymbol{\ell}\in\mathbb{R}^D$ is sampled independently from a centered Laplace distribution with scale $\frac{2}{\epsilon}$\\
    Adversarial prediction: $\hat{z}^i \leftarrow A(\textbf{x}^i_{priv})$ \\
    Update $\theta_A$ by backpropagating the loss $L(z^i,\hat{z}^i)$\\
    Task classification: $\hat{y}^i \leftarrow C(\textbf{x}^i_{priv})$ \\
    Update $\theta_E$ and $\theta_C$ by backpropagating the loss $L(y^i,\hat{y}^i) - \lambda \cdot L(z^i,\hat{z}^i)$
    }
\caption{Training procedure of \AdvNoise (one epoch).}
\label{alg:algo}
\end{algorithm*}

\section{Proof of Theorem~\ref{thm:priv}}
\label{app:privacy_proof}

\begin{proof}
We start by proving that our noisy encoder $E_{priv}:X\rightarrow\mathbb{R}^D$ satisfies $\epsilon$-DP. Recall that for any input text $x\in X$
$$E_{priv}(x)=priv\circ E(x)=E(x)/\|E(x)\|_1 + \boldsymbol{\ell},$$
where each entry of $\boldsymbol{\ell}\in\mathbb{R}^D$ is sampled independently from $\text{Lap}(\frac{2}{\epsilon})$, the centered Laplace distribution with scale $2/\epsilon$.
% where $E$ is the encoder and $priv$ adds Laplace noise with  the noise addition component.
Let $\tilde{E}(x)=E(x)/\|E(x)\|_1$. The L1 sensitivity of $\tilde{E}$ is
\begin{equation*}
    \Delta_{\tilde{E}} = \underset{x, x'\in X}{\max}~ % \sum_{i=1}^{D} |x_{i} - x'_{i}|.
    \|\tilde{E}(x) - \tilde{E}(x)'\|_1.
\end{equation*}
Since for any $x\in X$ we have $\|\tilde{E}(x)\|_1=1$, the triangle inequality gives $\Delta_{\tilde{E}}\leq 2$. The $\epsilon$-DP guarantee then follows from the application of the Laplace mechanism \cite{DBLP:conf/tcc/DworkMNS06}. Formally, let
$$p(y) = \frac{\epsilon}{4}e^{-\frac{|y|\epsilon}{2}}$$
denote the p.d.f. of $\text{Lap}(2/\epsilon)$. Consider two arbitrary input texts $x,x'\in X$ and let $\tilde{\textbf{x}}=\tilde{E}(x)\in\mathbb{R}^D$ and $\tilde{\textbf{x}}'=\tilde{E}(x')\in\mathbb{R}^D$ be their normalized encoded representations. Then, for any possible encoded output $\textbf{e}=(e_1,\dots,e_D)\in\mathbb{R}^D$, we have:
\begin{align}
    \frac{\Pr[E_{priv}(x)=\textbf{e}]}{\Pr[E_{priv}(x')=\textbf{e}]} &= \prod_{d=1}^D\frac{p(e_d - \tilde{x_d})}{p(e_d - \tilde{x_d}')}\label{eq:indep}\\
    &= \prod_{d=1}^D\frac{e^{-\frac{\epsilon}{2}|e_d - \tilde{x}_d|}}{e^{-\frac{\epsilon}{2}|e_d - \tilde{x}'_d|}}\nonumber\\
    &= e^{\frac{\epsilon}{2}\sum_{d=1}^D|e_d - \tilde{x}'_d| - |e_d - \tilde{x}_d|}\nonumber\\
    &\leq e^{\frac{\epsilon}{2}\sum_{d=1}^D|\tilde{x}_d- \tilde{x}'_d|}\label{eq:triangle}\\
    &= e^{\frac{\epsilon}{2}\|\tilde{\textbf{x}}- \tilde{\textbf{x}}'\|_1}\nonumber\\
    &\leq e^{\frac{\epsilon}{2}\Delta_{\tilde{E}}} = e^\epsilon,\label{eq:sensitivity}
\end{align}
where \eqref{eq:indep} follows from the independence of the noise across dimensions, \eqref{eq:triangle} uses the triangle inequality, and \eqref{eq:sensitivity} from the definition of $\Delta_{\tilde{E}}$ and the fact that $\Delta_{\tilde{E}}\leq 2$ as shown above.

The above inequality shows that $E_{priv}$ satisfies $\epsilon$-DP as per Definition~\ref{def:dp}.
The fact that $C\circ E_{priv}$ also satisfies $\epsilon$-DP follows from the post-processing property of DP, which ensures that the composition of any function with an $\epsilon$-DP algorithm also satisfies $\epsilon$-DP \cite{DBLP:journals/fttcs/DworkR14}.

\end{proof}

\section{Error in Privacy Analysis of Previous Work}
\label{sec:correction}

% of the representation is defined as: 
% \begin{equation}
%     \bigtriangleup f = \underset{\textbf{x}, \textbf{x'}}{\max} \sum_{i=1}^{D} |x_{i} - x'_{i}|
% \end{equation}

As briefly mentioned in Section~\ref{sec:related-work}, we found a critical error in the differential privacy analysis made in previous work by \citet{DBLP:conf/emnlp/LyuHL20}. This error is then reproduced in subsequent work by \citet{plant2021cape}. In this section, we explain this error and its consequences for the formal privacy guarantees of these methods, and provide a correction.

Recall from Section~\ref{sec-prelims} that to achieve $\epsilon$-DP with the Laplace mechanism, one must calibrate the scale of the Laplace noise needed to the L1 sensitivity of the encoded representation (see Eq.~\ref{eq:sens}). This sensitivity bounds the worst-case change in L1 norm for any two arbitrary encoded user inputs $\textbf{x}$ and $\textbf{x}'$ of dimension $D$.

In order to bound the L1 sensitivity, \citet{DBLP:conf/emnlp/LyuHL20} and \citet{plant2021cape} propose to bound each entry of the encoded input $\textbf{x}\in\mathbb{R}^D$ in the $[0,1]$ range. Specifically, they normalize as follows:
% \begin{equation}
%     \textbf{x} = \textbf{x} - \textbf{x}_{min} / \textbf{x}_{max} - \textbf{x} 
% \end{equation}
\begin{align}
\label{eq:norm_lyu}
    &\textbf{x} \leftarrow \textbf{x} - \min(\textbf{x}) / (\max(\textbf{x}) - \min(\textbf{x})),
%    & \textbf{x}_{max} = \max_{\textbf{x}} x_i \forall i \in D \\
%    & \textbf{x}_{min} = \min_{\textbf{x}} x_i \forall i \in D
\end{align}
where $\min(\textbf{x})$ and $\max(\textbf{x})$ are respectively the minimum and maximum values in the vector $\textbf{x}$. \citet{DBLP:conf/emnlp/LyuHL20} and \citet{plant2021cape} incorrectly claim that this allows to bound the L1 sensitivity by $1$ and thus add Laplace noise of scale $\frac{1}{\epsilon}$ . In fact, the sensitivity can be as large as $D$, as can be seen by considering the two inputs $\textbf{x}=[0, 1,\dots, 1]_D$ and $\textbf{x}'=[1, 0, \dots, 0]$ for which 
$\|\textbf{x}-\textbf{x}'\|_1=D$. Therefore, to achieve $\epsilon$-DP, the scale of the Laplace noise should be $\frac{D}{\epsilon}$ (i.e., $D$ times larger than what the authors use). As a consequence, the differential privacy provided by their method are $D$ times worse than claimed by \citet{DBLP:conf/emnlp/LyuHL20} and \citet{plant2021cape}: the $\epsilon$ values they report should be multiplied by $D$, which leads to essentially void privacy guarantees.

% Another more subtle problem with the normalization in Eq.~\ref{eq:norm_lyu} is that the range $[0,1]$ should hold for all possible text representations $\textbf{x}$ (not only the ones in the observed dataset but also any future point seen at inference time) for the DP guarantees to hold. However, computing $\textbf{x}_{min}$ and $\textbf{x}_{max}$ may not be sufficient to guarantee this, as it is possible that there exists an input text $x$ whose encoded representation $\textbf{x}$ would lead to larger or smaller values than previously observed.

While \citet{DBLP:conf/emnlp/LyuHL20} claim to follow the approach of \citet{priv_deep_learning}, they missed the fact that \citet{priv_deep_learning} do account for multiple dimensions by scaling the noise to the number of entries (denoted by $c$ in their paper) that are submitted to the server, see pseudo-code in Figure~12 of \citet{priv_deep_learning}.

In contrast to \citet{DBLP:conf/emnlp/LyuHL20} and \citet{plant2021cape}, our normalization in Eq.~\ref{eq:our_lpriv} guarantees by design that the L1 sensitivity is bounded by $2$. We provide a complete and self-contained proof of our privacy guarantees in Section~\ref{app:privacy_proof}.

% are the smallest and the largest value of the component of $x$ respectively. With the proposed construction, for the input of D dimension, the sensitivity should be D. As one can construct two inputs $\textbf{x}=[1,1, ..., 1]_D$ and $\textbf{x'}=[0,0, ..., 0]_D$ for which the L1-sensitivity is D. However, they incorrectly assume the sensitivity to be 1 and adds noise proportional to $Lap(\frac{1}{\epsilon})$ instead of $Lap(\frac{D}{\epsilon})$. This results in significantly less privacy guarantees of their approach then they claim.

\section{Experiments}
This section gives more information on the experimental setup and also provides additional results.
% We begin by defining FPR-gap.

% \subsection{Fairness Measure}
% \label{sec:fpr}

% \paragraph{FPR-gap:} Formally, for a classifier $C$, with binary labels $y\in\{0,1\}$ and protected attribute $z\in\{g,\neg g\}$
% , FPR-gap is defined as:
% \begin{equation}
%         \textrm{FPR-gap} = P_g(\hat{Y}=0|Y=0) - P_{\neg g}(\hat{Y}=0|Y=0)
% \end{equation}
% where $\hat{Y}$ is the predicted class.

\subsection{Privacy metric}
\label{sec:appendix-privacy-metric}

\paragraph{Leakage:} We compute the leakage using a sklearn’s MLPClassifier. We use the validation set of the original dataset as the train and the test set of the original dataset as the test.

\paragraph{Minimum Description Length (MDL)} is a information-theoretic probing measure which captures the strength of regularity in the data. In this work, we employ the online coding approach \citep{DBLP:conf/emnlp/VoitaT20} to calculate MDL. Online coding captures the regularity by characterizing the effort required to achieve a certain level of accuracy. Here, a portion of data is transmitted to the receiver at each step, which then uses all the data in the previous steps to understand the regularity in the current step. The regularity is obtained by training the model on the previously received data and then evaluating it on the current portion of the data.

Borrowing, the terminology from~\citet{DBLP:conf/emnlp/VoitaT20}, consider a dataset $D$ consisting of  $\{ (x_1,y_1), \cdots, (x_n,y_n)  \}$ pairs, where the $x_i$'s are the data representation, and the $y_i$'s are the task label. In our case, $x_i$ is the output of the encoder, and $y_i$ is the sensitive attribute associated with the underlying text. Following the standard information theory setting, consider a sender Alice who wants to transmit labels  $y_{1:n}=\{y_1 \cdots, y_n \}$ to a receiver Bob, and both of them have access to the data representation $x_{1:n}=\{x_1 \cdots, x_n \}$. In order to transmit labels $y_{1:n}$ efficiently (as few bits possible), Alice encodes $y_{1:n}$ using a model $p(y|x)$. According to Shannon-Huffman code, the minimum bits required to transmit these labels losslessly is:

$$L_p (y_{1:n} | x_{1:n}) = - \sum^{n}_{i=1} \log_2 p(y_i | x_i).$$

In the online coding setting of MDL, the labels are transmitted in blocks of n timesteps $t_0 < t_1 < \cdots t_n$. Alice starts by encoding $y_{1:t_1}$ with a uniform code, then
both Alice and Bob learn a model $p_{\theta_1} (y|x)$ that predicts y from x using data $\{(x_i, y_i)\}^{t1}_{i=1}$.  Alice then uses this model to communicate the next data block $y_{t_1:t_2}$, and both learns a new model using larger chunk of data  $\{(x_i, y_i)\}^{t2}_{i=1}$. This continues till the whole set of labels $y_{1:n}$ is transmitted. The total code length required for transmission using this setting is given as:
\begin{equation} \label{eq1}
\begin{split}
L_{online}(y_{1:n} | x_{1:n}) & = t_{1}\log_{2}C - \\
 & \sum^{n-1}_{i=1} \log_{2}p_{\theta_i}(y_{t_i+1:t_i}|x_{t_i+1:t_i} ).
\end{split}
\end{equation}
where $y_{i} \in \{1,2, \cdots, C\}$. In our case, the online code length $L_{online}(y_{1:n} | x_{1:n})$ is shorter, if it is easier for probing model to perform well with fewer training instances. This implies that the sensitive information is more easily available in the encoder's representation.

We compute MDL using sklearn’s MLPClassifier at timesteps corresponding to $0.1\%$, $0.2\%$, $0.4\%$, $0.8\%$, $1.6\%$, $3.2\%$, $6.25\%$, $12.5\%$, $25\%$, $50\%$ and $100\%$ of each dataset as suggested by~\citet{DBLP:conf/emnlp/VoitaT20}.

\subsection{Datasets}
\label{appendix-dataset}
\paragraph{Twitter Sentiment} \citep{blodgett-etal-2016-demographic} consists of 200k tweets annotated with a binary sentiment label and a binary ``race'' attribute corresponding to African American English (AAE) vs. Standard American English (SAE) speakers. The initial representation of tweets are obtained from a Deepmoji encoder~\cite{DBLP:conf/emnlp/FelboMSRL17}.
The dataset is evenly balanced with respect to the four sentiment-race subgroup combinations. To create bias in the training data, we follow \citet{DBLP:conf/emnlp/ElazarG18} and change the race proportion in each sentiment class to have 40\% AAE-happy, 10\% AAE-sad, 10\% SAE-happy, and 40\% SAE-sad. Test data remains balanced.
This setup is particularly challenging regarding privacy and fairness, as the model may exploit the correlation between the protected attribute and the main class label, which is reinforced due to skewing. 
% It is further compounded by the fact that the underlying language used in the dataset itself is very predictive of the protected attribute~\cite{DBLP:conf/emnlp/ElazarG18}.
The mismatch between the train-test distribution is also relevant for our setup, where the system may be trained on publicly available datasets or collected via an opt-in policy and may therefore not closely resemble the test distribution. This dataset is made available for research purposes only.\footnote{\url{http://slanglab.cs.umass.edu/TwitterAAE/}}

\paragraph{Bias in Bios} \cite{DBLP:conf/fat/De-ArteagaRWCBC19} consists of 393,423 textual biographies
% which were collected by scrapping the web and are
annotated with an occupation label (28 classes) and a binary gender attribute. Similar to~\citet{DBLP:conf/acl/RavfogelEGTG20}, we encode each biography with BERT~\cite{DBLP:conf/naacl/DevlinCLT19}, using the last hidden state over the CLS token. We use the same train-valid-test split as~\citet{DBLP:conf/fat/De-ArteagaRWCBC19}. As the dataset was collected by scrapping the web, it tends to reflect common gender stereotypes and contains explicit gender indicators (e.g., pronouns), making it more challenging to prevent models from relying on these gendered words. It is also more complex than Twitter Sentiment in terms of the number of classes. Dataset is released under MIT License.\footnote{\url{https://github.com/Microsoft/biosbias}}

\paragraph{CelebA}~\cite{DBLP:conf/iccv/LiuLWT15} consists of over 200,000 images of the human face, alongside with 40 binary attributes labels describing the content of the images. Following the standard setting as described in~\cite{pmlr-v119-lohaus20a}, we use 38 of these attributes as features, "Smiling" as the class label, and "Sex" as the sensitive attribute. We use 60\% of the data as train, 20\% as validation, and the remaining as the test split. The CelebA dataset is available for non-commercial research purposes.\footnote{\url{https://mmlab.ie.cuhk.edu.hk/projects/CelebA.html}}

\paragraph{Adult Income}~\cite{DBLP:conf/kdd/Kohavi96} consists of a U.S. 1994 Census database segment and has 48842 instances with 14 features each. We apply the pre-processing as proposed by~\cite{DBLP:conf/www/Wu0W19} resulting in a total of 9 features for each instance. The objective is to predict whether a given data point earns more than fifty thousand U.S. dollars or less. We consider sex (binary) as the sensitive attribute. Like CelebA,  We use 60\% of the data as train, 20\% as validation, and the remaining as the test split. The license of the dataset is unknown, however it is commonly used in several fairness papers and is avaialbe at~\cite{Dua:2019}. 

% \paragraph{Twitter Sentiment}~\cite{DBLP:conf/fat/De-ArteagaRWCBC19} This dataset is evenly balanced with respect to the four sentiment-race subgroup combinations. To create bias in the training data, we follow \citet{DBLP:conf/emnlp/ElazarG18} and change the race proportion in each sentiment class to have 40\% AAE-happy, 10\% AAE-sad, 10\% SAE-happy, and 40\% SAE-sad. Test data remains balanced.
% This setup is particularly challenging regarding privacy and fairness, as the model may exploit the correlation between the protected attribute and the main class label, which is reinforced due to skewing. 
% It is further compounded by the fact that the underlying language used in the dataset itself is very predictive of the protected attribute~\cite{DBLP:conf/emnlp/ElazarG18}.

\subsection{Model Architecture}
\label{sec:architecture}

\paragraph{Twitter Sentiment.} The encoder consists of two layers with ReLU activation and a fixed dropout of 0.1. The classifier is linear, and the adversarial branch consists of three layers. We use a fixed dropout of 0.1 in all the layers with ReLU activation, apart from the last layer. 

\paragraph{Bias in Bios.} The encoder consists of three layers and a fixed dropout of 0.1. The classifier also consists of three layers, and the adversarial branch consists of two layers. We use a fixed dropout of 0.1 in all the layers with ReLU activation, apart from the last layer. 

In case of \textit{Adult Income} and \textit{CelebA} dataset we use the same model as for \textit{Twitter Sentiment}.

\begin{table*}[t]
\centering
\adjustbox{max width=\linewidth}{
\begin{tabular}{lrrrr}
\toprule
 Method              &   Accuracy $\uparrow$ &     TPR-gap $\downarrow$ &  Leakage $\downarrow$   &  MDL $\uparrow$\\
\midrule
\texttt{Random} & 50.00 $\pm$ 0.00 &   0.00 $\pm$ 0.00 &   --   &  104.64 $\pm$ 0.11 \\
 \Uncon            & 85.70 $\pm$ 0.21 & 12.25 $\pm$ 2.07  & 81.3 $\pm$ 0.89 & 67.82 $\pm$ 1.46\\
 \midrule \midrule
 \texttt{INLP}         &     84.81  $\pm$ 0.47 & 12.69 $\pm$ 4.66 &  66.00 $\pm$ 1.32 & 100.17 $\pm$ 1.65\\
 \texttt{Noise}               & 85.12 $\pm$ 0.47 & 12.49 $\pm$ 0.58 & 59.01 $\pm$ 0.65 & 103.93 $\pm$ 0.24\\
 \Adv         & 85.34 $\pm$ 0.22  & 7.83 $\pm$ 0.97 & 87.00 $\pm$ 2.22 & 46.61 $\pm$ 5.52\\
 \AdvMultiple & 84.92 $\pm$ 0.12 & 5.79 $\pm$ 1.44 & 84.38 $\pm$ 2.07 & 51.11 $\pm$ 4.06\\
 \midrule \midrule
 \AdvNoise & 84.81 $\pm$ 0.34 & 2.68 $\pm$ 0.60 & 65.49 $\pm$ 3.48 & 98.53 $\pm$ 4.51 \\
\bottomrule
\end{tabular}}
\caption{\label{tab:celeb-results-main} Test results on CelebA dataset with fixed Relaxation Threshold of 1.0. Fairness is measured by TPR-Gap (lower is better), while privacy is measured by Leakage (lower is better) and MDL (higher is better). The MDL achieved by \texttt{Random} gives an upper bound for that particular dataset. The results have been averaged over $5$ different seeds.}
\end{table*}

\begin{table*}[t]
\centering
\adjustbox{max width=\linewidth}{
\begin{tabular}{lrrrr}
\toprule
 Method              &   Accuracy $\uparrow$ &     TPR-gap $\downarrow$ &  Leakage $\downarrow$   &  MDL $\uparrow$\\
\midrule
\texttt{Random} & 50.00 $\pm$ 0.00 &   0.00 $\pm$ 0.00 &   -   &   20.15 $\pm$ 0.083\\
 \Uncon             & 83.41 $\pm$ 0.32 & 12.73 $\pm$ 7.17 & 78.19 $\pm$ 1.0  & 16.38 $\pm$ 0.46\\
 \midrule \midrule
 \texttt{INLP}         &     83.11  $\pm$ 0.51 & 3.91 $\pm$ 2.43 &  74.54 $\pm$ 0.67 & 19.93 $\pm$ 0.35\\
 \texttt{Noise}               & 82.87 $\pm$ 0.37 & 8.01 $\pm$ 1.18  & 68.12 $\pm$ 0.94 & 19.38 $\pm$ 0.33 \\
 \Adv         & 83.14 $\pm$ 0.53 & 7.02 $\pm$ 3.31  & 78.2 $\pm$ 0.18  & 16.1 $\pm$ 0.36 \\
 \AdvMultiple & 83.14 $\pm$ 0.25 & 3.55 $\pm$ 2.16  & 81.37 $\pm$ 0.98 & 13.5 $\pm$ 1.09\\
 \midrule \midrule
 \AdvNoise & 82.29 $\pm$ 0.9  & 2.73 $\pm$ 2.18  & 70.25 $\pm$ 4.81 & 18.1 $\pm$ 2.79 \\
\bottomrule
\end{tabular}}
\caption{\label{tab:adult-results-main} Test results on Adult Income dataset with fixed Relaxation Threshold of 1.0. Fairness is measured by TPR-Gap (lower is better), while privacy is measured by Leakage (lower is better) and MDL (higher is better). The MDL achieved by \texttt{Random} gives an upper bound for that particular dataset. The results have been averaged over $5$ different seeds.}
\end{table*}

\subsection{Hyperparameters}
\label{sec:hyper}

For all our experiments, we use Adam optimizer with a learning rate of 0.001 and batch size of 2000. We give additional tuning details of the different methods below. A single experiment takes about 30 minutes to run on Intel Xenon CPU. We will also provide the PyTorch model description in the README of the source code for easier reproduction.

\begin{itemize}
    \item \Adv:
    % In adversarial learning, the adversary is generally connected to the encoder via a gradient reversal layer which scales the loss with $-\lambda$ during backward pass. In this work, as a part of the hyperparameter tuning,
    We perform a grid search over $\lambda$ varying it between $0.1$ to $3.0$ with an interval of $0.2$. Moreover, following previous work~\cite{DBLP:conf/nips/LampleZUBDR17,DBLP:conf/icassp/AdiZCULS19}, instead of a constant $\lambda$, we increase it over the epochs using the update scheme $\lambda_i = 2/(1+e^{-p_i}) -1$, where $p_i$ is the scaled version of the epoch number. We also experimented with increasing the $\lambda$ linearly, as well as keeping it constant, but found the above update scheme to perform the best in various settings. We also use this scheme in all other adversarial approaches. 
    \item \AdvMultiple: Similar to \Adv, we vary $\lambda$  between $0.1$ to $3.0$ with an interval of $0.2$. Apart from $\lambda$, \AdvMultiple has an additional hyperparameter $\lambda_{ort}$ which corrresponds to the weight given to the orthogonality loss component. We vary $\lambda_{ort}$ between $0.1$ and $1.0$. Here, we do a simultaneous grid search over $\lambda$ and $\lambda_{ort}$ resulting in 150 runs for each seed. We fix the number of the adversary to three which is the same as the original implementation by~\cite{DBLP:conf/eacl/HanBC21}.
    \item \AdvNoise: In order to have comparable number of runs to \AdvMultiple, we experiments with following $\epsilon$ values: $8.0, 9.0, 10.0, 11.0, 12.0, 13.0, 14.0, 15.0, 16.0,$ $20.0$. Similar to above approach, we do a simultaneous grid search over $\lambda$ and $\epsilon$ resulting in 150 runs for each seed.
    \item \texttt{INLP}: In the case of \texttt{INLP}, we always debias the representation after the penultimate classifier layer and before the final layer, which is consistent with the setting considered by the authors \citep{DBLP:conf/acl/RavfogelEGTG20}. We also observe that this choice empirically led to the best results. We vary the number of iterations as a part of hyperparameter tuning. For Bias in Bios we vary the iterations between 15 and 45, while for Twitter Sentiment we vary between 2 to 7. We found that in case of Bias in Bios, performing less than 15 iterations resulted in the same behaviour as \Uncon model over validation set while more than 45 iterations resulted in a random classifier. We observed the same in the Twitter Sentiment before 2 and after 7 iterations, respectively.
\end{itemize}

\subsection{Extended Evaluation}
\label{sec-appendix-extended-evaluation}
Tables~\ref{tab:celeb-results-main}\nobreakdash--\ref{tab:adult-results-main} present detailed results on CelebA and Adult Income dataset respectively. In terms of fairness over both the datasets, we observe that adversarial-based approaches induce a more fair model than Unconstrained or Noise, with \AdvNoise outperforming all other methods.  Interestingly, unlike Twitter Sentiment and Bias in Bios, all approaches have comparable accuracy, including \texttt{Noise} and \texttt{INLP}. We believe this to be the case due to these datasets being relatively more challenging than CelebA and Adult Income. As observed previously, purely adversarial-based approaches leak significantly more information than the DP-based approaches in terms of privacy.  We observe that \texttt{Noise} and \texttt{INLP} performs marginally better in privacy than \AdvNoise; however, they suffer significantly in the fairness metric. In fact, they induce fairness levels which are similar to \Uncon.

Overall, the results show \AdvNoise as the only viable choice to induce a fairer model and make its representation private while maintaining comparable accuracy. These observations are in line with previous experiments described in Sec.~\ref{sec:relaxation}

\subsection{Additional Results}
\label{sec:appendix-results}

Tables~\ref{tab:encoded-emoji-results-appendix-1}\nobreakdash--\ref{tab:encoded-emoji-results-appendix-3} present detailed results on Twitter Sentiment with different relaxation thresholds, which were summarized in Figure~\ref{fig:rt-tradeoff}.

Table~\ref{tab:encoded-emoji-results-appendix-4} provides the detailed privacy-fairness results which were summarized in Figure~\ref{fig:accuracy-privacy-graph}.

\begin{center}
\begin{table*}[t]
\centering
\adjustbox{max width=\linewidth}{
\begin{tabular}{lrrr}
\toprule
 Method                &     Accuracy $\uparrow$ &   TPR-gap $\downarrow$ &     Leakage $\downarrow$ \\
\midrule
\Uncon & 72.54 $\pm$ 0.57 & 27.17 $\pm$ 1.76 & 87.18 $\pm$ 0.32  \\
\midrule \midrule
\texttt{Noise} & 71.87 $\pm$ 0.56 & 25.14 $\pm$ 3.47 &  71.75 $\pm$ 2.99  \\
\Adv        & 75.49 $\pm$ 0.71 & 8.47 $\pm$ 3.5   &  88.03 $\pm$ 0.24   \\
 \AdvMultiple        & 75.6 $\pm$ 0.53  & 7.74 $\pm$ 4.17  &  88.01 $\pm$ 0.28 \\
 
 \midrule \midrule
 
 \AdvNoise & 75.34 $\pm$ 0.56 & 5.46 $\pm$ 3.59  &  62.31 $\pm$ 5.69 \\
 
%  Adversarial + Differentiated + Noise & 74.61 &      1.79 &     17.44 &     62.1  &           54.42 \\
\bottomrule
\end{tabular}}
\caption{\label{tab:encoded-emoji-results-appendix-1} Test set results on Twitter Sentiment dataset (scores averaged over 5 different seeds, RT=$0.0$).}
\end{table*}
\end{center}

\begin{center}
\begin{table*}[t]
\centering
\adjustbox{max width=\linewidth}{
\begin{tabular}{lrrrr}
\toprule
 Method                &     Accuracy $\uparrow$ &   TPR-gap $\downarrow$ &    Leakage $\downarrow$ \\
\midrule
\Uncon & 70.57 $\pm$ 0.98 & 20.68 $\pm$ 0.99 &  82.91 $\pm$ 1.65  \\
\midrule \midrule
\texttt{Noise} & 70.47 $\pm$ 0.43 & 19.84 $\pm$ 0.91 &  66.83 $\pm$ 3.32   \\
\Adv        & 74.09 $\pm$ 1.56 & 3.03 $\pm$ 2.65  & 88.14 $\pm$ 0.18  \\
 \AdvMultiple       & 74.44 $\pm$ 0.62 & 1.07 $\pm$ 0.74  &  87.98 $\pm$ 0.36  \\
 
 \midrule \midrule
 
 \AdvNoise & 74.24 $\pm$ 1.25 & 0.89 $\pm$ 0.46  &  61.92 $\pm$ 5.04  \\
 
%  Adversarial + Differentiated + Noise & 74.61 &      1.79 &     17.44 &     62.1  &           54.42 \\
\bottomrule
\end{tabular}}
\caption{\label{tab:encoded-emoji-results-appendix-2} Test set results on Twitter Sentiment dataset (scores averaged over 5 different seeds, RT=$3.0$).}
\end{table*}
\end{center}

\begin{center}
\begin{table*}[t]
\centering
\adjustbox{max width=\linewidth}{
\begin{tabular}{lrrrrr}
\toprule
 Method                &     Accuracy $\uparrow$ &   TPR-gap $\downarrow$ &    Leakage $\downarrow$\\
\midrule
\Uncon  & 70.57 $\pm$ 0.98 & 20.68 $\pm$ 0.99 &  82.91 $\pm$ 1.65    \\
\midrule \midrule
\texttt{Noise} & 70.47 $\pm$ 0.43 & 19.84 $\pm$ 0.91 &  66.83 $\pm$ 3.32   \\
\Adv        & 70.8 $\pm$ 2.77  & 1.72 $\pm$ 1.5   &  88.2 $\pm$ 0.24    \\
 \AdvMultiple       & 67.39 $\pm$ 1.16 & 1.0 $\pm$ 0.8    &  88.01 $\pm$ 0.12    \\
 
 \midrule \midrule
 
 \AdvNoise & 73.97 $\pm$ 1.6  & 1.4 $\pm$ 1.22   & 60.38 $\pm$ 5.46   \\
 
%  Adversarial + Differentiated + Noise & 74.61 &      1.79 &     17.44 &     62.1  &           54.42 \\
\bottomrule
\end{tabular}}
\caption{\label{tab:encoded-emoji-results-appendix-3} Test set results on Twitter Sentiment dataset (scores averaged over 5 different seeds, RT=$10.0$).}
\end{table*}
\end{center}

\begin{center}
\begin{table*}[t]
\centering
\adjustbox{max width=\linewidth}{
\begin{tabular}{@{}rrrrrr@{}}
\toprule
\multirow{2}{*}{Method} & \multirow{2}{*}{$\epsilon$} & \multicolumn{2}{c}{Twitter Sentiment} & \multicolumn{2}{c}{Bias in Bios} \\
                        &                      & Accuracy  $\uparrow$        & Leakage   $\downarrow$       & Accuracy  $\uparrow$      & Leakage  $\downarrow$       \\ \midrule \midrule
\texttt{Noise}                   & 8.0                  & 71.3  &   60.59               & 64.75 &    56             \\
\AdvNoise     & 8.0                  & 74.89 &   56.91               & 64.78 &    54.4             \\ \midrule \midrule
\texttt{Noise}                   & 10.0                 & 71.63 &   65.57               & 70.86 &    57.7             \\
\AdvNoise      & 10.0                 & 75.25 &   60.55               & 70.97 &    56.5             \\ \midrule \midrule
\texttt{Noise}                   & 12.0                 & 71.76 &   66.04               & 75.01 &    58.4             \\
\AdvNoise     & 12.0                 & 75.31 &   53.31               & 75.01 &    57             \\ \midrule \midrule
\texttt{Noise}                   & 14.0                 & 71.7  &   67.98               & 76.74 &    59             \\
\AdvNoise     & 14.0                 & 75.3  &   57.29               & 76.83 &    56.3             \\ \midrule \midrule
\texttt{Noise}                   & 16.0                 & 71.7  &   67.69               & 77.77 &    60.3             \\
\AdvNoise     & 16.0                 & 75.56 &   61.98               & 77.89 &    57.9             \\
\bottomrule
\end{tabular}

}
\caption{\label{tab:encoded-emoji-results-appendix-4} Accuracy-privacy trade-off for different noise level (as captured by $\epsilon$).}
\end{table*}
\end{center}

\end{document}